\documentclass{article}

\usepackage[preprint]{neurips_2021}

\usepackage[utf8]{inputenc}
\usepackage[T1]{fontenc}
\usepackage{hyperref}
\usepackage{url}
\usepackage{booktabs}
\usepackage{amsfonts}
\usepackage{nicefrac}
\usepackage{microtype}
\usepackage{xcolor}
\usepackage{graphicx}
\usepackage{float}
\usepackage{amsmath}
\usepackage{caption}
\usepackage{subcaption}
\usepackage{array}
\usepackage{tabularx}

\definecolor{darkblue}{rgb}{0,0.08,0.45}
\hypersetup{
    colorlinks=true,
    linkcolor=darkblue,
    citecolor=darkblue,
    filecolor=darkblue,
    urlcolor=darkblue,
}

\title{OstrichRL: A Musculoskeletal Ostrich Simulation to Study Bio-mechanical Locomotion}

\author{
Vittorio La Barbera\thanks{Equal contribution. The models, tasks and data are available at \url{https://github.com/vittorione94/ostrichrl} and visualizations at \url{https://sites.google.com/view/ostrichrl}.} ~$^{1}$ \hspace{9pt} Fabio Pardo$^{*2}$ \hspace{9pt} Yuval Tassa$^{3}$ \hspace{9pt} Monica A. Daley$^{4}$ \\
\bf Christopher T. Richards$^{1}$ \hspace{9pt} Petar Kormushev$^{2}$ \hspace{9pt} John R. Hutchinson$^{1}$ \vspace{9pt} \\
\small{$^{1}$Royal Veterinary College, London, $^{2}$Imperial College London} \\
\small{$^{3}$DeepMind, London, $^{4}$University of California, Irvine} \vspace{4pt} \\
{\tt\small \{vlabarbera, jhutchinson, ctrichards\}@rvc.ac.uk,} \\
{\tt\small \{f.pardo, p.kormushev\}@imperial.ac.uk,} \\
{\tt\small tassa@deepmind.com, madaley@uci.edu}
}

\begin{document}
\textit{}
\maketitle

\begin{abstract}
Muscle-actuated control is a research topic that spans multiple domains, including biomechanics, neuroscience, reinforcement learning, robotics, and graphics. This type of control is particularly challenging as bodies are often overactuated and dynamics are delayed and non-linear. It is however a very well tested and tuned actuation mechanism that has undergone millions of years of evolution with interesting properties exploiting passive forces and efficient energy storage of muscle-tendon units. To facilitate research on muscle-actuated simulation, we release a 3D musculoskeletal simulation of an ostrich based on the MuJoCo physics engine. The ostrich is one of the fastest bipeds on earth and therefore makes an excellent model for studying muscle-actuated bipedal locomotion. The model is based on CT scans and dissections used to collect actual muscle data, such as insertion sites, lengths, and pennation angles. Along with this model, we also provide a set of reinforcement learning tasks, including reference motion tracking, running, and neck control, used to infer muscle actuation patterns. The reference motion data is based on motion capture clips of various behaviors that we preprocessed and adapted to our model. This paper describes how the model was built and iteratively improved using the tasks. We also evaluate the accuracy of the muscle actuation patterns by comparing them to experimentally collected electromyographic data from locomoting birds. The results demonstrate the need for rich reward signals or regularization techniques to constrain muscle excitations and produce realistic movements. Overall, we believe that this work can provide a useful bridge between fields of research interested in muscle actuation.
\end{abstract}

\begin{figure}[t]
\centering
\includegraphics[width=2.3cm]{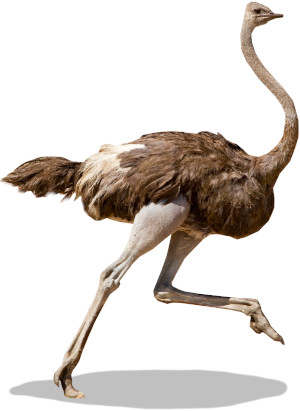}
\hspace{0.12cm}
\hfill
\includegraphics[width=2.1cm]{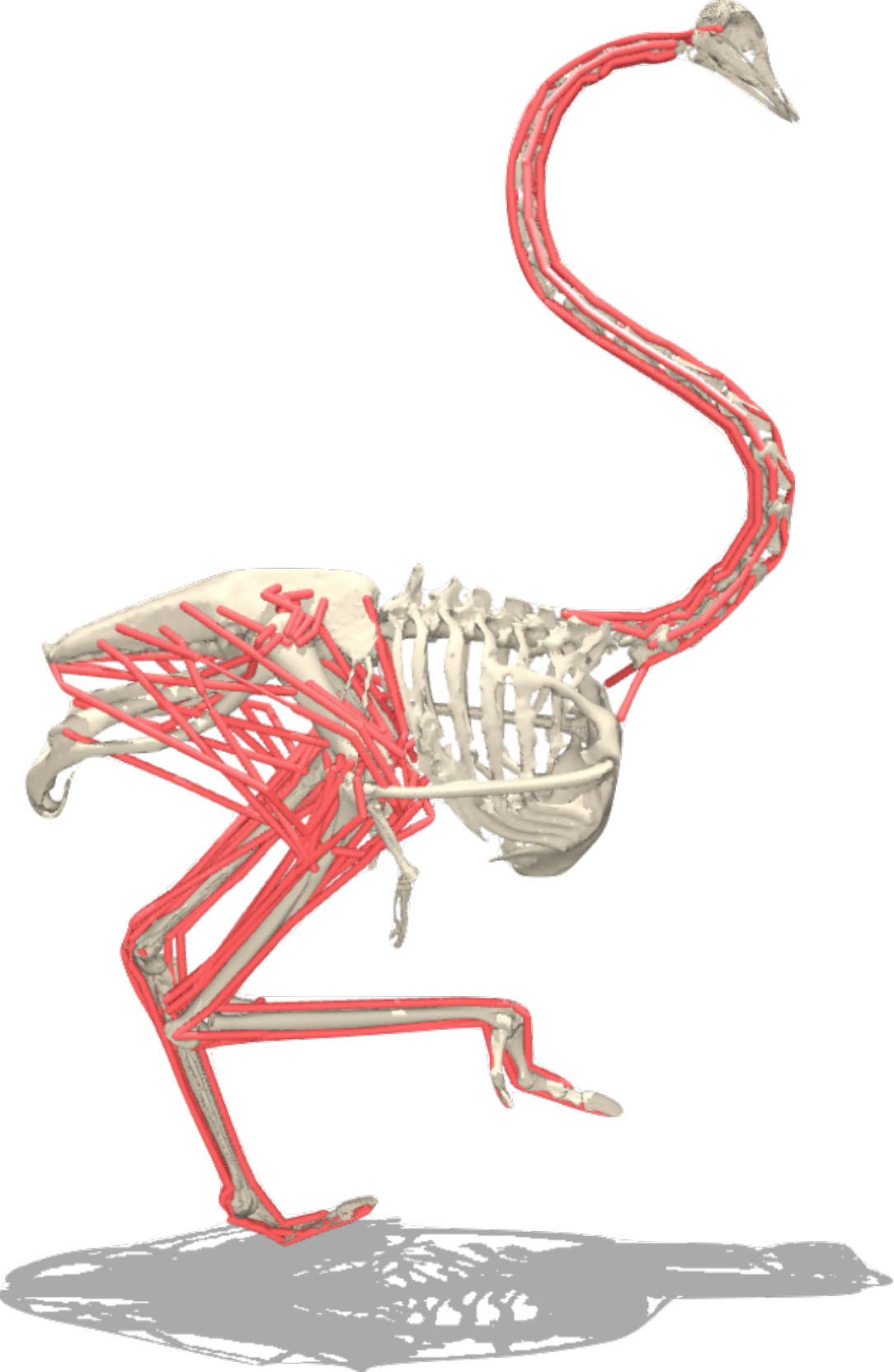}
\hfill
\includegraphics[width=2.1cm]{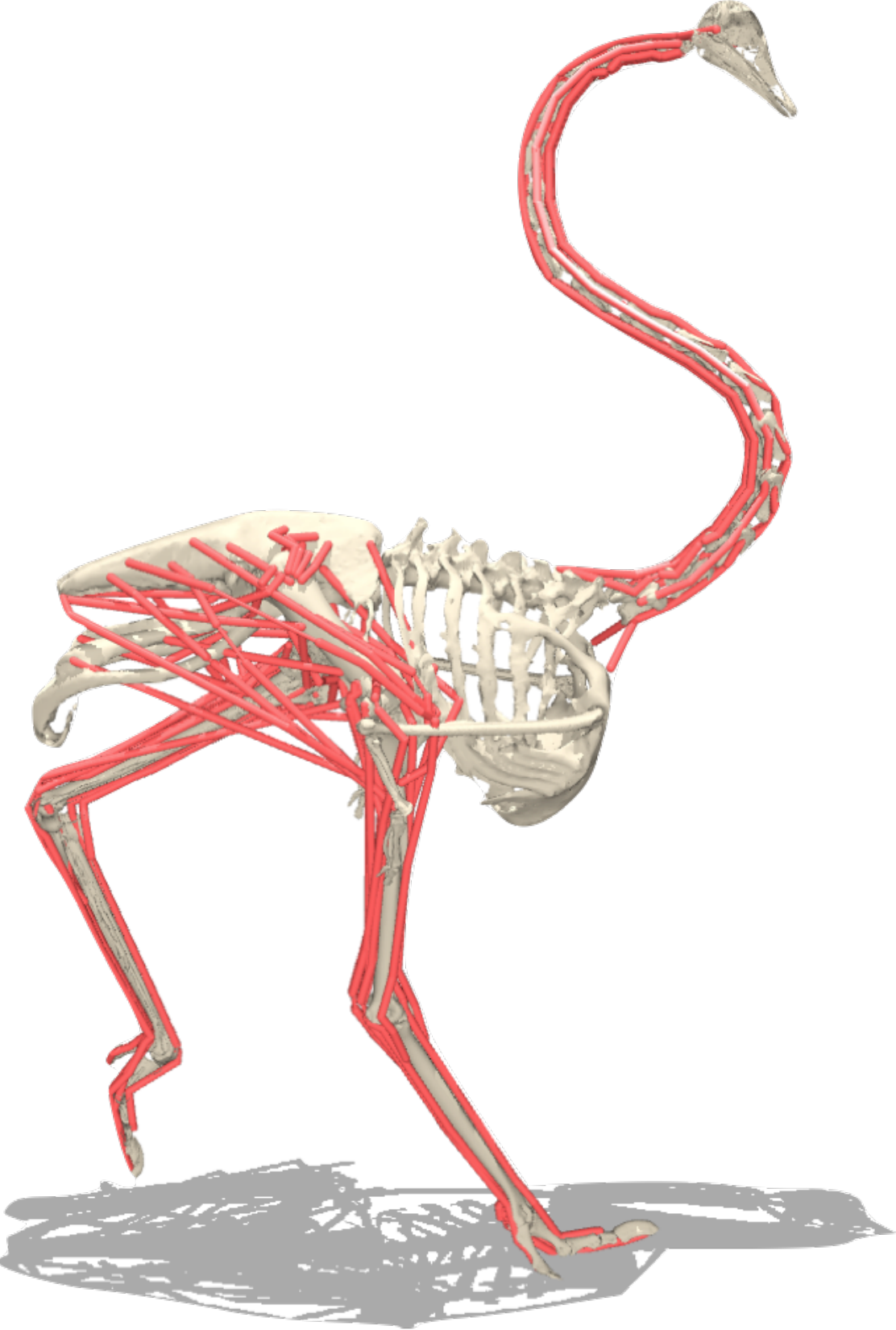}
\hfill
\includegraphics[width=2.1cm]{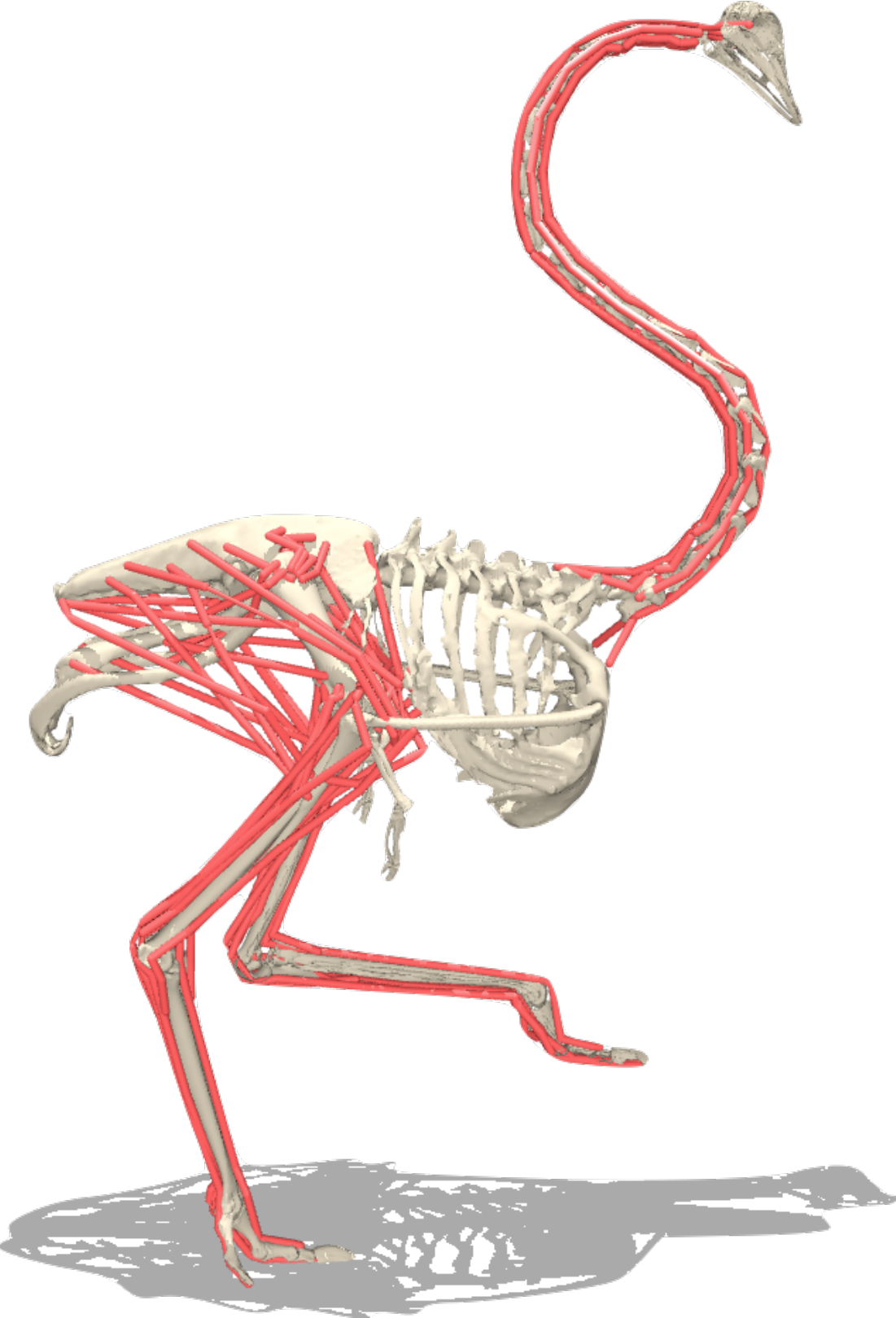}
\hfill
\includegraphics[width=2.1cm]{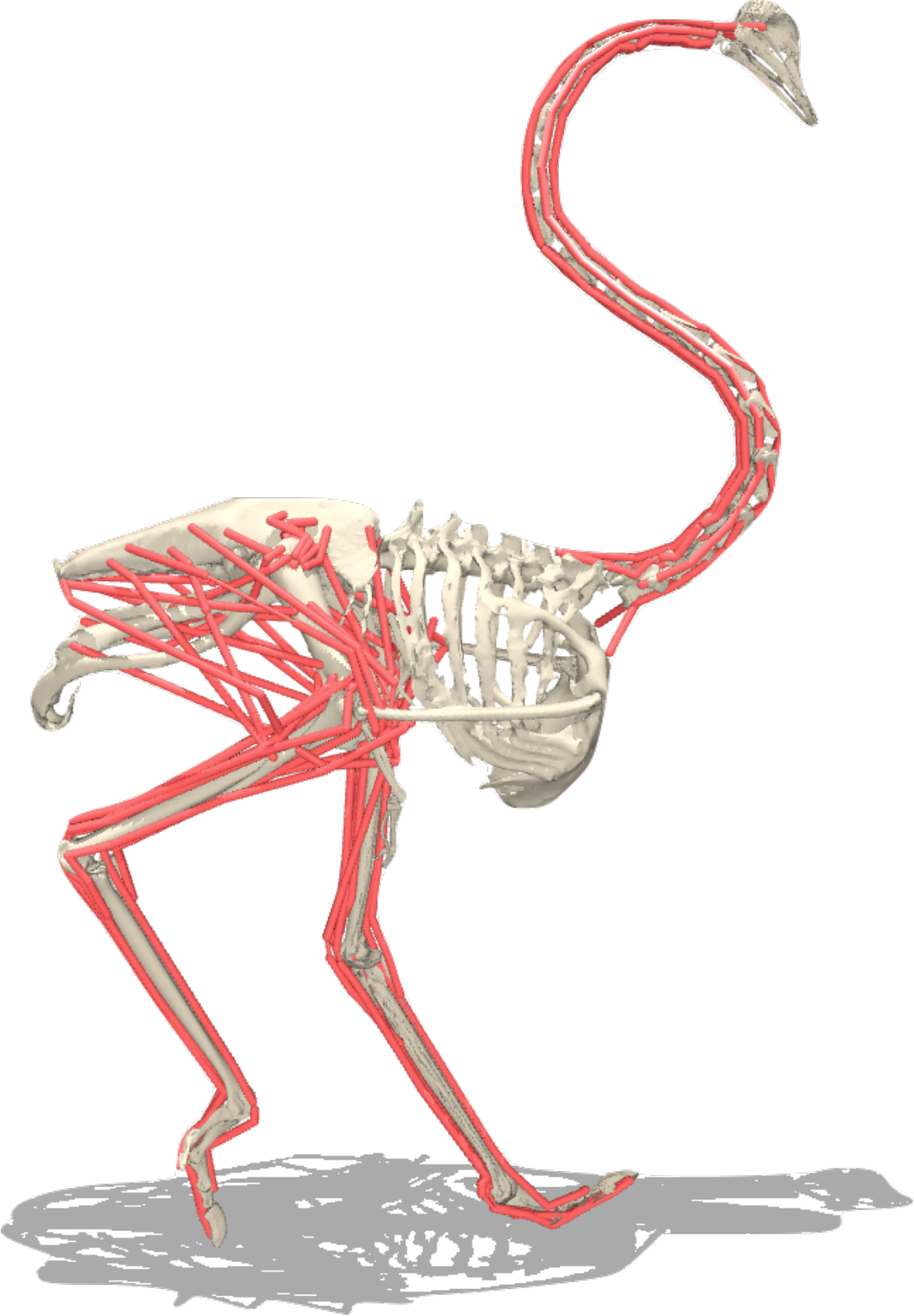}
\hfill
\includegraphics[width=1.8cm]{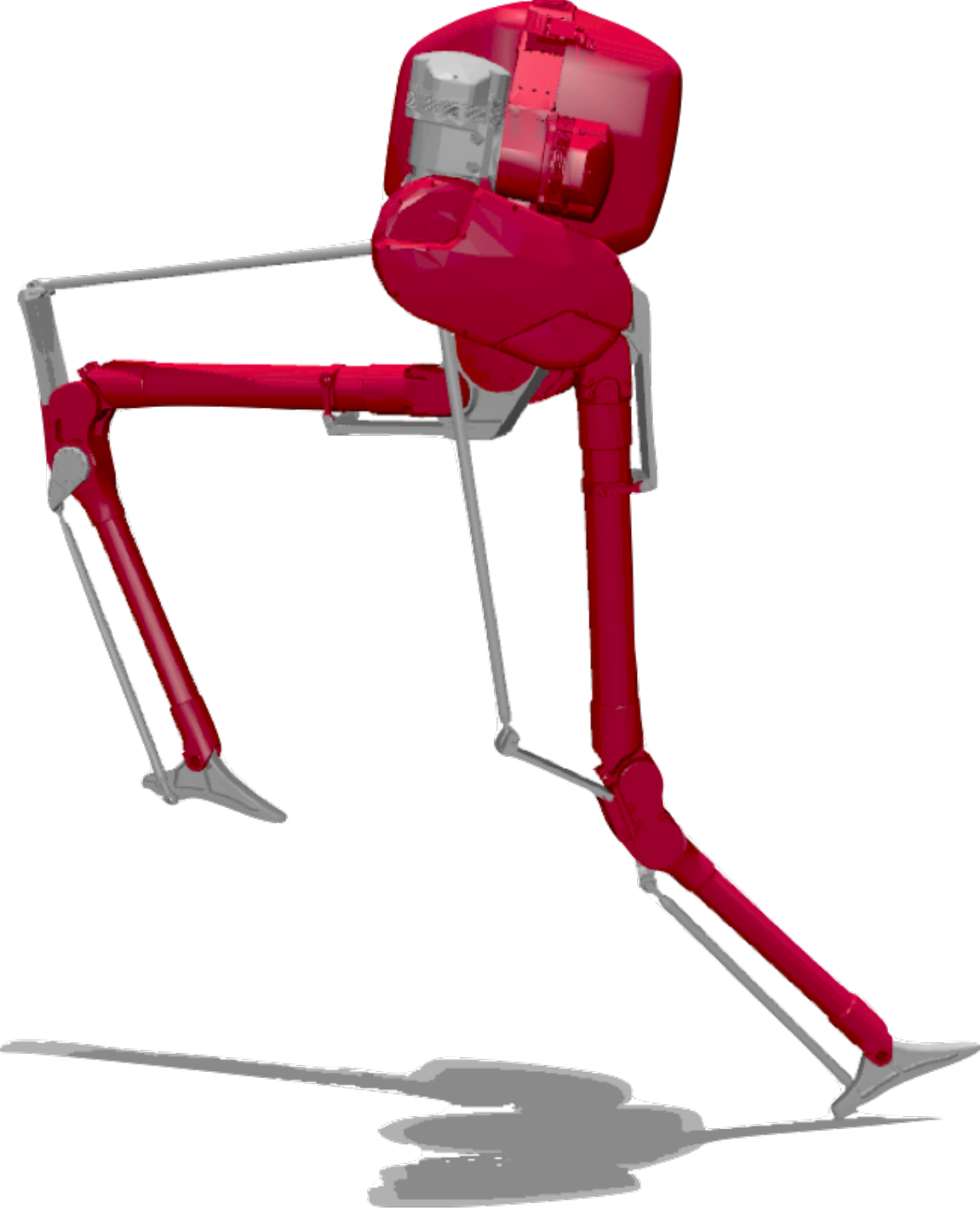}
\caption[Comparison between an ostrich, the proposed model, and Cassie]{The proposed musculoskeletal model is based on data collected from real ostriches, while the Cassie robot provides a robotic approximation of the morphology of an ostrich.}
\label{fig:ostrichrl_illustration}
\end{figure}

\section{Introduction}

In nature, vertebrates produce movement by the contraction of skeletal muscles that pull on the bones, creating torque at the joints. Understanding muscles is of interest in many different fields. First, from a biological point of view, it is desirable to understand how animals use their bodies to perform various behaviors. Muscles are also relevant in computer graphics to obtain accurate skin deformations in virtual characters and to produce more natural-looking gaits for films and video games \citep{angles2019viper,abdrashitov2021interactive,modi2021emu}. In sports medicine, musculoskeletal simulations can be used to understand sports injuries \citep{bulat2019musculoskeletal} or create effective training and recovery strategies \citep{coste2017comparison}. Unfortunately, in robotics, with the exception of soft robots, muscles have been studied relatively little despite their interesting energetic properties and the sophisticated movements they enable \citep{wang2021recent,cotton2012fastrunner}.

Biomechanics is the study of how biological systems produce motion. Biological motion is affected by many factors such as the musculoskeletal geometry, internal states of the muscles, and force-length-velocity curves. Understanding how complex biological systems such as humans and other animals use their muscles to move is quite challenging. Obtaining data from living animals is difficult and not neutral to animal welfare. To overcome this problem, biomechanics researchers often use computer-based simulations combined with numerical optimization to estimate how muscles are used. However, most of the research that uses optimization techniques does not leverage recent progress made in reinforcement learning.

In this work, we propose a new ostrich musculoskeletal model, illustrated in \autoref{fig:ostrichrl_illustration}, that uses a fast physics engine with reinforcement learning tasks to infer muscle actuation patterns. Ostriches are interesting to study because they are fast, economical bipedal runners \citep{alexander1979mechanics}. To the best of our knowledge, we are the first to apply RL to such a realistic animal model, solving locomotion tasks. With the proposed open-source computational tools, we hope to open new opportunities for researchers interested in accurate and fast muscle simulation provided by MuJoCo combined with RL.

\section{Related work}

\subsection{Building musculoskeletal models}

The NeurIPS conference has been hosting recurrent competitions to bridge the gap between biomechanics and reinforcement learning
\footnote{\url{https://www.crowdai.org/challenges/nips-2017-learning-to-run} \\ \indent\hspace{0.17cm} \url{https://www.crowdai.org/challenges/nips-2018-ai-for-prosthetics-challenge} \\ \indent\hspace{0.17cm} \url{https://www.aicrowd.com/challenges/neurips-2019-learning-to-move-walk-around}}
, where the goal was to make a musculoskeletal human model walk. The challenge used the OpenSim \citep{delp2007opensim,seth2018opensim} simulator. However, as documented in many of the solutions \citep{pavlov2018run,zhou2019efficient,akimov2019distributed,kolesnikov2020sample}, the engine was too slow to properly use traditional deep RL methods, which often require millions or billions of samples to find good solutions. Moreover, the model used in the latest challenge was in many ways quite simplistic, actuating only the legs with 22 muscles in total and removing the arms.

Simulating musculoskeletal animal models is not common outside the field of biomechanics. Most of the efforts are focused on humans because of available resources in anatomical atlases and interest from the entertainment industry and sports. Available full-body animal musculoskeletal models include, for example the chimpanzee \citep{sellers2013exploring} and dog \citep{stark2021three}. Moreover, for human models, researchers tend to have access to much higher quality motion capture data than for animals, but animal research often have superior empirical measurements of physiological function.

\subsection{Controlling musculoskeletal models}

Some innovative research has used deep RL to solve complex locomotion tasks such as playing basketball \citep{liu2018learning}, performing athletic jumps \citep{yin2021discovering}, or boxing and fencing \citep{won2021control}. These studies used motion capture data to build a rich reward signal during training \citep{merel2017learning,merel2018neural,hasenclever2020comic}. \citet{wang2012optimizing} was one of the first studies in the graphics community to use biological actuators. \citet{geijtenbeek2013flexible} used evolution-based algorithms to control muscular bipeds and also optimized muscle routing. \citet{jiang2019synthesis} tried to bridge torque-actuated models and muscle dynamics, using neural networks to map muscle activations to torque commands and achieved natural looking gaits. \citet{lee2018dexterous} used volumetric muscles and trajectory optimization for juggling. \citet{lee2019scalable} combined RL and motion capture clips to control a human musculoskeletal model solving a variety of tasks from walking to weight lifting, separating muscle coordination from trajectory mimicking through two different networks with privileged information and an intermediate proportional derivative (PD) controller.

\section{Methods}

\begin{figure}[t]
\centering
\includegraphics[width=\linewidth]{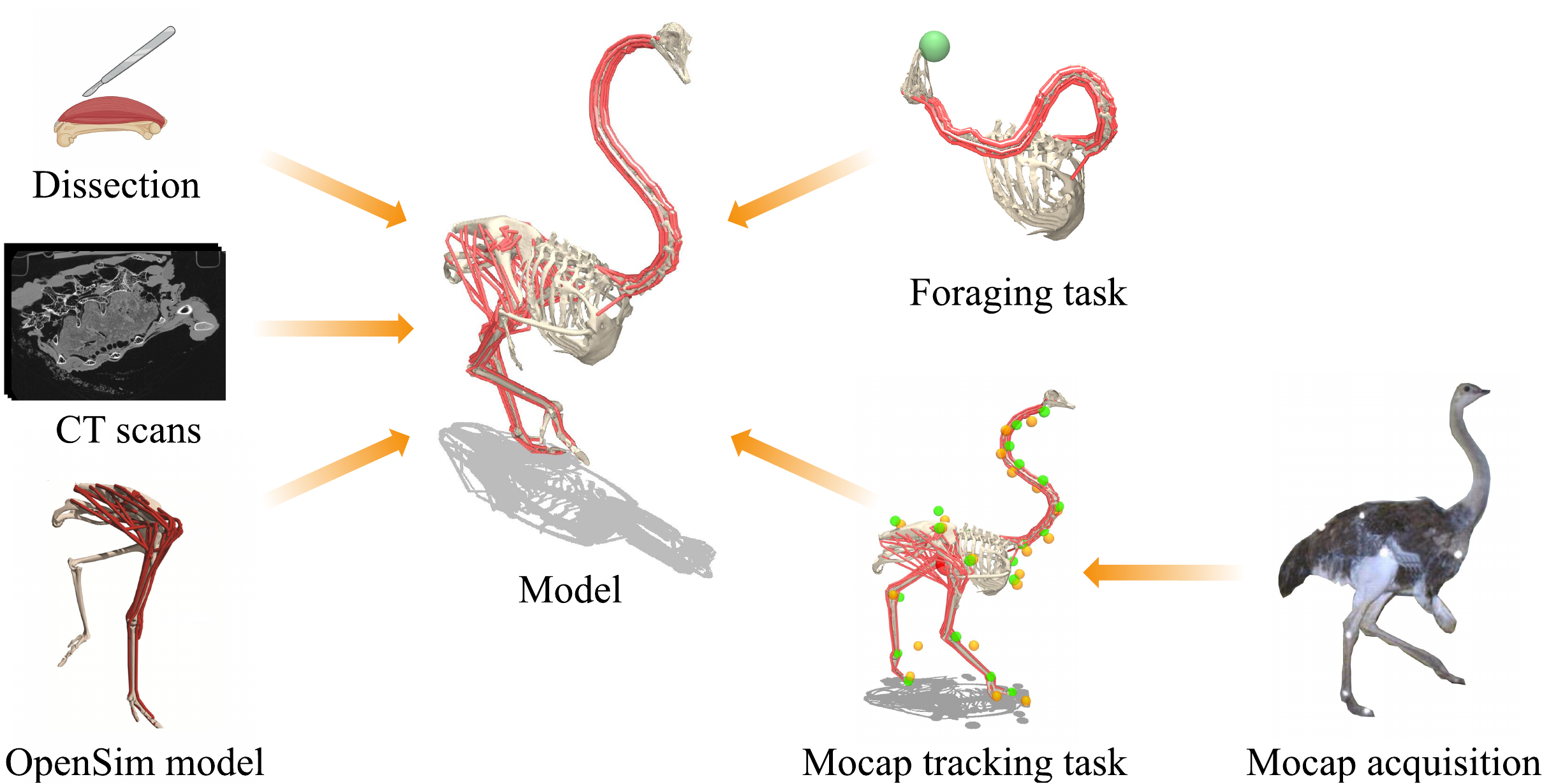}
\caption[Workflow used to build the ostrich model]{The workflow used to build the ostrich model. Various sources of data were blended together for the modeling part, and tasks helped in fine-tuning muscle strength and joint limits.}
\label{fig:ostrichrl_workflow}
\end{figure}

This paper describes how a fast and realistic musculoskeletal model of an ostrich was built by combining several sources of information, including an existing OpenSim model of legs, a dissection, a CT scan, and reinforcement learning tasks that allowed us to iteratively improve properties of the model. This workflow is represented in \autoref{fig:ostrichrl_workflow}.

\subsection{Anatomical data acquisition}

\paragraph{Computed tomography scanning}
We first acquired multiple computed tomography (CT) scans of a $96.5$ kg adult male ostrich specimen, donated by a local farm. The CT scans were then segmented (separating the bone from the rest of the x-ray slices) using the Mimics software (Materialise, Inc.) to obtain 3D models (polygonal meshes) for each bone from the ribcage to the head, including the wings. These 3D geometries were important for two reasons: firstly, they provided an accurate representation of the geometry of the bones, vital for defining muscle attachment points, and secondly, they allowed inference of the inertia of body segments. To obtain the inertia of body segments we followed the steps provided in \citet{hutchinson20073d}. The idea is to wrap the skeleton parts within meshes representing flesh and assume constant density. From there, volumes, masses, and the inertia tensors were computed.

\paragraph{Muscle dissection}
Next, we performed a dissection in order to gather data for muscles of the neck and rib cage. Wings are not actuated in our model because they are mostly used when turning at high velocity and air friction is not considered in our simulations. Dissection is important to properly understand muscle routing, where the muscles start (origin) and where they end (insertion). During the dissection, we used anatomical descriptions from \citet{bohmer2019gulper}, \citet{tsuihiji2005homologies}, and \citet{tsuihiji2007homologies} to identify the different muscles. We also acquired muscle-specific data: muscle mass, pennation angle, tendon length, and muscle fiber lengths, used when simulating the muscle-tendon dynamics.

\subsection{Modeling}

\paragraph{OpenSim and MuJoCo}
Two computer modeling and simulation software packages are commonly used by the biomechanics and reinforcement learning communities respectively. The first one is OpenSim \citep{delp2007opensim,seth2018opensim}, an open source engine that provides benchmarked physics-based muscle simulations. It is relevant to our study not only for its popularity, but also because the legs of our ostrich model were originally modeled with this engine \citep{hutchinson2015musculoskeletal}. The same model was also used to run some tracking (inverse) simulations to estimate the forces, activations, and mechanical work of the muscles involved in solving locomotion tasks \citep{rankin2016inferring}. OpenSim is rather slow in simulating each time step, as documented in previous NeurIPS competitions that used it to simulate muscle dynamics \citep{kolesnikov2020sample}. The second simulator, MuJoCo \citep{todorov2012mujoco}, has recently been open-sourced, and is used in multiple RL domains, because it provides fast and accurate rigid body dynamics. Specifically, tendon routing in OpenSim uses an iterative algorithm, while MuJoCo's tendon routing is closed-form and, therefore, much faster. In a comparison made by \citet{ikkala2020converting}, when calculating the average run time over 97 forward simulations, MuJoCo was roughly 600 times faster than OpenSim.

\paragraph{Musculoskeletal models}
Both OpenSim and MuJoCo define their models in a hierarchical tree structure. Muscles are attached to at least two bones, using two sites at the extremities, called the origin and the insertion sites. Other sites, called waypoints, and wrapping geometries, can also be specified to define elaborate muscle routings. Waypoints are sites through which the muscle must pass, which are useful to maintain an anatomically realistic 3D path for the muscles. Wrapping geometries are useful to prevent muscles from penetrating the bones, and are geometric primitives such as spheres or cylinders that muscles must wrap. 

\begin{figure}[t]
\centering
\includegraphics[height=5cm]{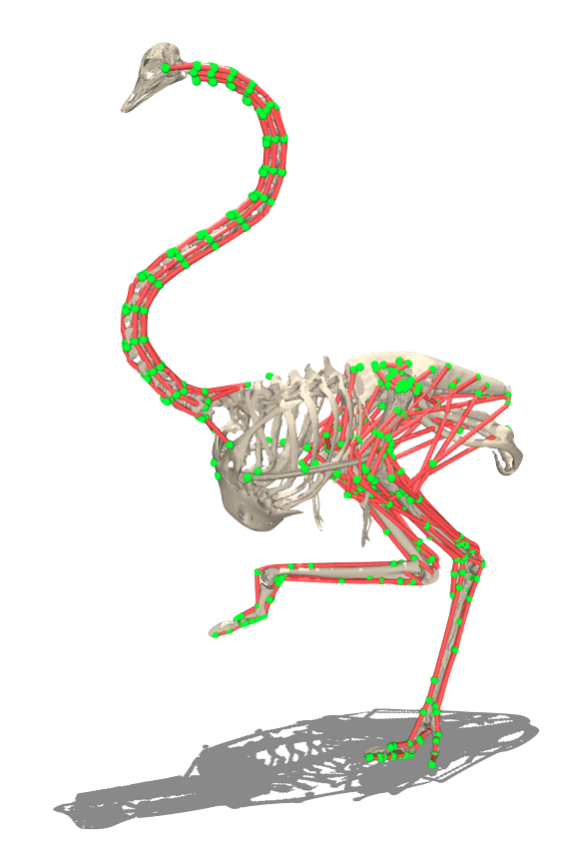}
\includegraphics[height=5cm]{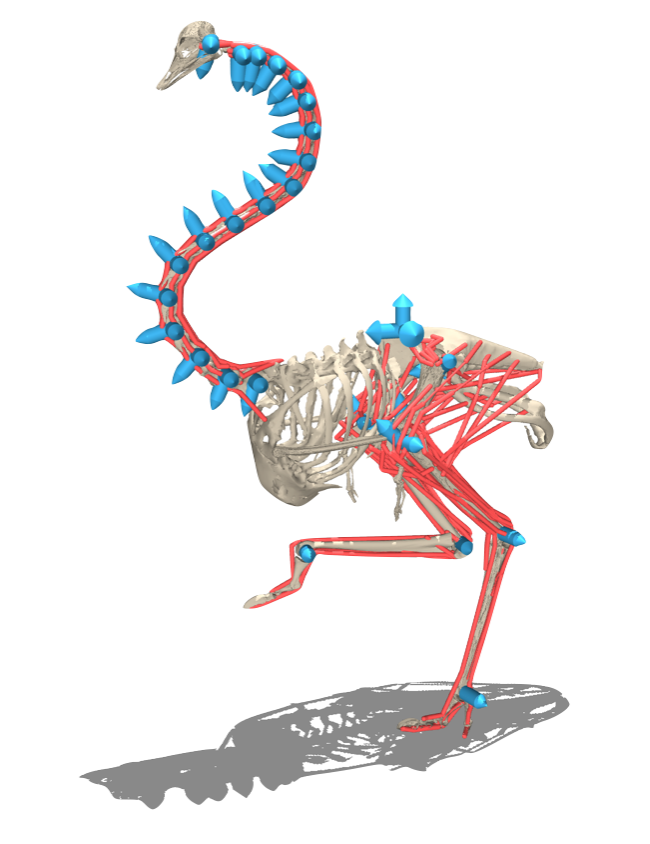}
\includegraphics[height=5cm]{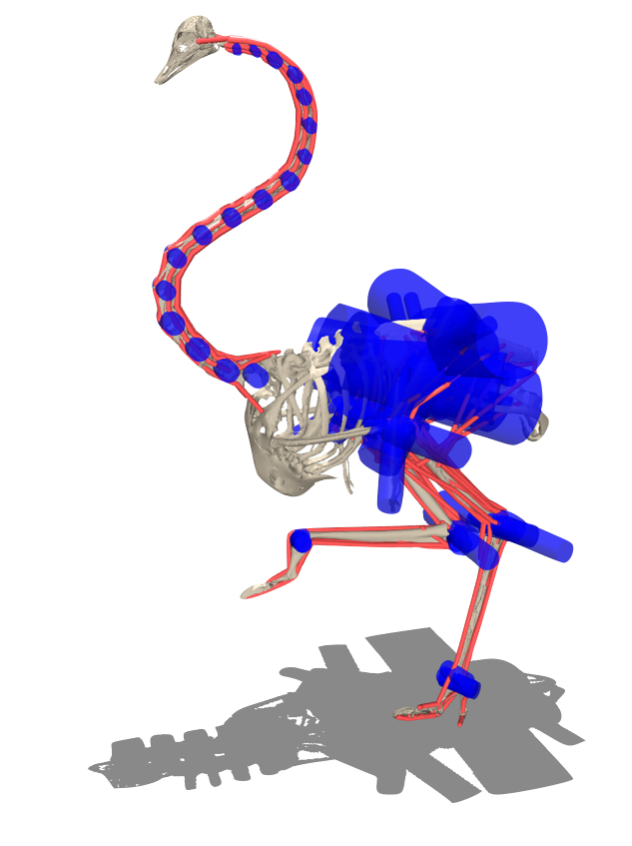}
\caption[Ostrich model]{The skeleton of the model follows a tree structure of bodies and joints. The muscle routing is defined using waypoints, shown in green on the left. The joints are shown in light blue in the middle. The wrapping geometries are shown in blue on the right (some wrapping geometries have been omitted for clarity).}
\label{fig:ostrichrl_model}
\end{figure}

\paragraph{Proposed model}
After converting the OpenSim leg model to MuJoCo's format, we completed the ostrich model by adding the ribcage, wings, neck, and head geometries. We then actuated the neck by adding the muscles, matching the routing from the dissection. We also modified the morphology of the feet, which were initially flat, to make their shape more realistic. We found that fine tuning the muscle lengths was necessary to improve stability. To find the new muscle ranges, we randomized the model joint angles in their limits a large number of times and recorded the global minimum and maximum lengths for each muscle. In our final model, visible in \autoref{fig:ostrichrl_model}, each leg contains 7 hinge joints: 3 for the hip, 2 for the knee, 1 for the ankle, and 1 for the metatarsophalangeal (mtp), while the neck contains 36 hinge joints, 2 for each pair of vertebrae and for the head. These 50 joints are actuated by 120 muscles, 68 in the legs and 52 in the neck, making this model very challenging to control. In comparison, the OpenSim model used for the NeurIPS 2019: Learn to Move - Walk Around challenge had only 22 muscles.

\paragraph{Simulation for RL}
MuJoCo has already been used in a number of popular control domains for reinforcement learning research, in particular dm\_control \citep{tassa2020dm_control} and OpenAI Gym \citep{brockman2016openai}. Most of these domains use simplistic models, often inspired by animal morphologies, made up of basic geometric bodies such as capsules, and torque-controlled multiaxial joints. While such models are fast and fairly easy to control, they do not provide a realistic basis for studying animal movement. On the other hand, simulations used in biomechanics typically model animals more accurately \citep{sellers2013exploring,hutchinson2015musculoskeletal,stark2021three} but are slow and often used in conjunction with relatively simple control techniques from the trajectory optimisation repertoire, such as direct collocation.

\subsection{Muscle dynamics}
\label{subsec:ostrichrl_muscle_dynamics}

\paragraph{Muscle excitation and activation}
In biomechanics and neuroscience, researchers separate muscle excitation and activation. A neural excitation $u$, produced by the nervous system, is responsible for the contraction of muscle fibers via an intermediate state called activation $a$ \citep{zajac1989muscle}. This intermediate state converts an electrochemical signal to mechanical force output. In MuJoCo, this conversion is modelled as a first-order nonlinear filter:
$$\frac{\partial a}{\partial t} = \frac{(u - a)}{\tau(u, a)}$$
where $\tau$ is defined as:
$$\tau(u, a) = \begin{cases} 
\tau_{a} (0.5 + 1.5 a) & u > a \\
\tau_{d} / (0.5 + 1.5 a) & u \leq a 
\end{cases}$$
$\tau_{a}$, $\tau_{d}$ are time constants for activation and deactivation, equal by default to 10 ms and 40 ms, respectively.

\begin{figure}[t]
\centering
\includegraphics[height=5.5cm]{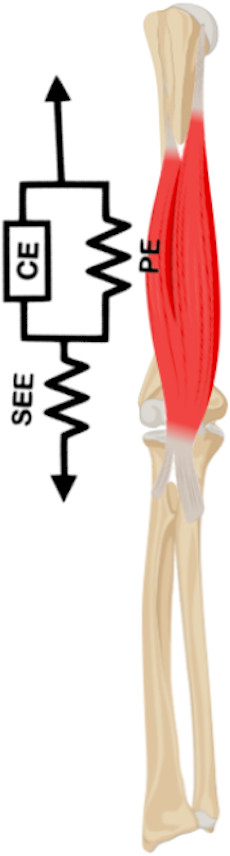}
\hspace{1cm}
\includegraphics[height=5.5cm]{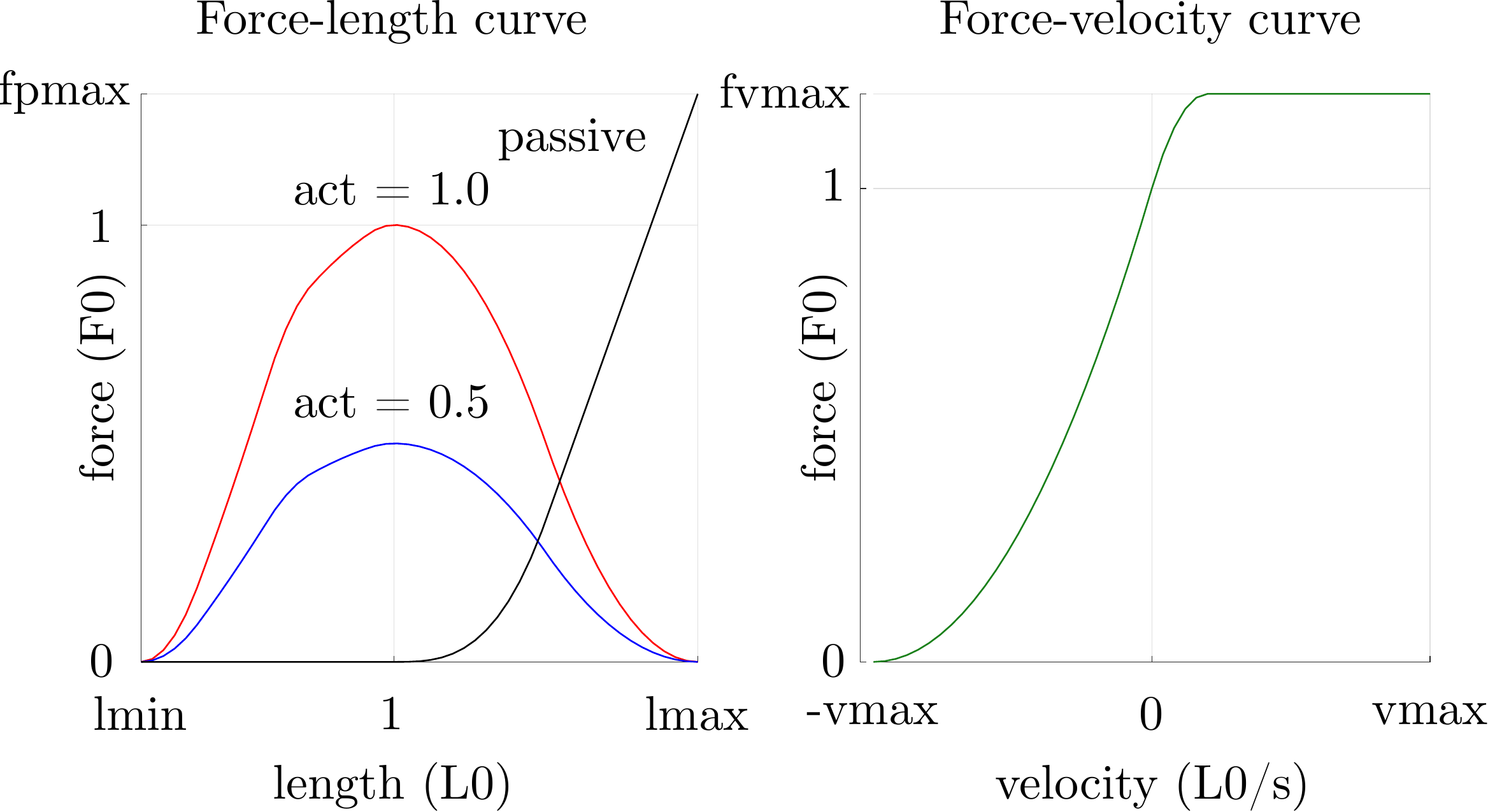}
\caption[Muscle dynamics]{On the left a diagram of the Hill-type muscle model comprising the 3 components: a contractile element, a passive element and an elastic element. On the right two plots showing the curves for the force-length and force-velocity functions used in the contraction dynamics.}
\label{fig:ostrichrl_contraction_dynamics}
\end{figure}

\paragraph{Contraction dynamics}
The contraction dynamics are implemented using the Hill-type model as shown in figure \autoref{fig:ostrichrl_contraction_dynamics}. This mechanical model computes the total force generated by a muscle by adding together the active contractile and passive properties of muscles. The contractile element  models the muscle contraction force, scaled by the activation signal a. The passive properties are modeled like springs: if a muscle is stretched, it will generate a passive force to return to its rest position.

The active contraction dynamics can be summarized by the formula:
$$f(l, \dot{l}, a) = a \cdot f_l(l) \cdot f_v(\dot{l}) + f_p(l)$$
where $f_l$ is the active force as a function of muscle length, $f_v$ is the active force as a function of velocity, and $f_p$ is the passive force which is always present regardless of activation. The force-length and force-velocity curves are showed in \autoref{fig:ostrichrl_contraction_dynamics}. These functions are built into MuJoCo, capturing the contraction dynamics for basic purposes, and have been validated by the biomechanics community for some conditions \citep{millard2013flexing}.

All muscle-related quantities in MuJoCo are scaled by the muscle-tendon actuator's resting length. The advantage of this representation is that all muscles behave similarly. MuJoCo does not allow specification of the muscle resting length $L_0$ and tendon slack length $LT$ directly, in contrast to OpenSim. This is due to the fact that MuJoCo does not include a stateful elastic component in the muscle model. This is a major compromise in MuJoCo when compared to OpenSim, that allows it to run faster. Instead, the actuator length range $LR$ needs to be specified, which is the minimum and maximum of the sum of muscle and tendon lengths, along with a range $R$ in units of $L_0$. In other words, the actuator length range $LR$ defines the interval of values that the actuator is allowed to use during the simulation and the range $R$ is a percentage expressing how much the actuator (here simply called ``muscle'') can shorten or extend.

At model compile time, MuJoCo solves these two equations where the unknowns are the muscle rest length $L_0$ and tendon length $LT$:
\begin{equation*}
\begin{aligned}
(LR_{min} - LT) / L_0 = R_{min} \\
(LR_{max} - LT) / L_0 = R_{max}
\end{aligned}
\end{equation*}

\subsection{Motion capture data}
\label{subsec:ostrichrl_mocap}

\begin{figure}[t]
\centering
\includegraphics[width=0.4\linewidth]{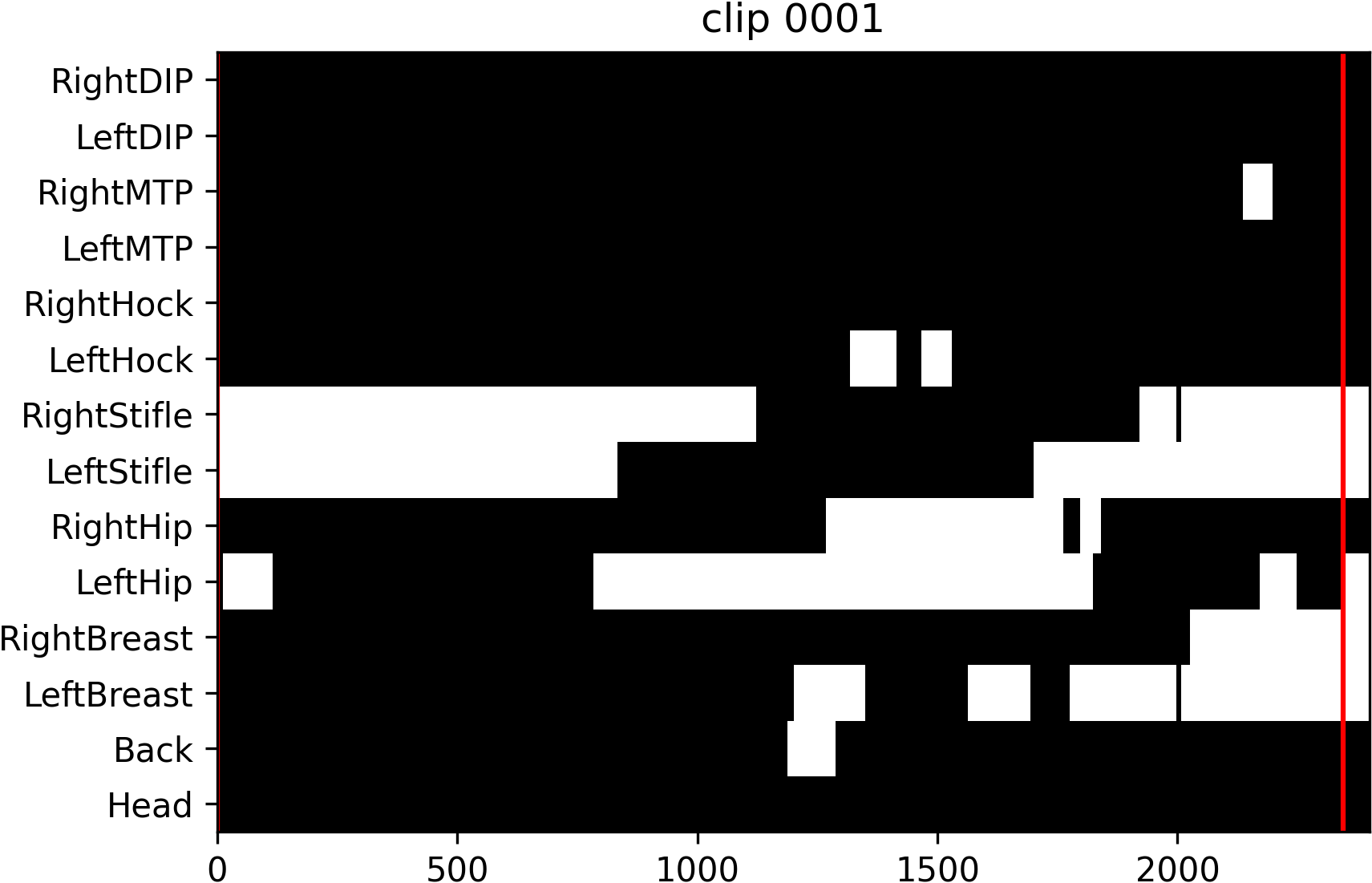}
\includegraphics[width=0.4\linewidth]{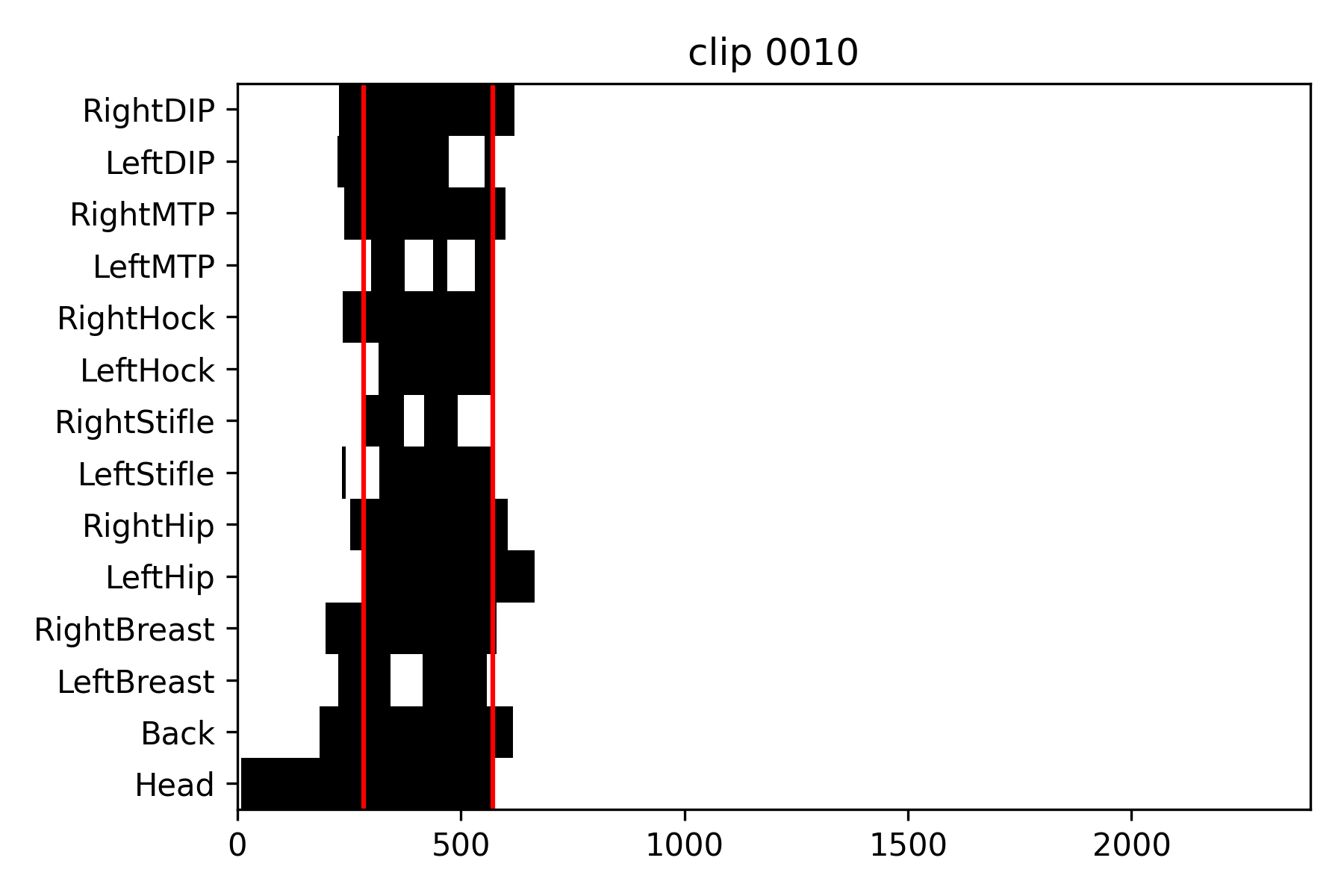} \\
\includegraphics[width=0.4\linewidth]{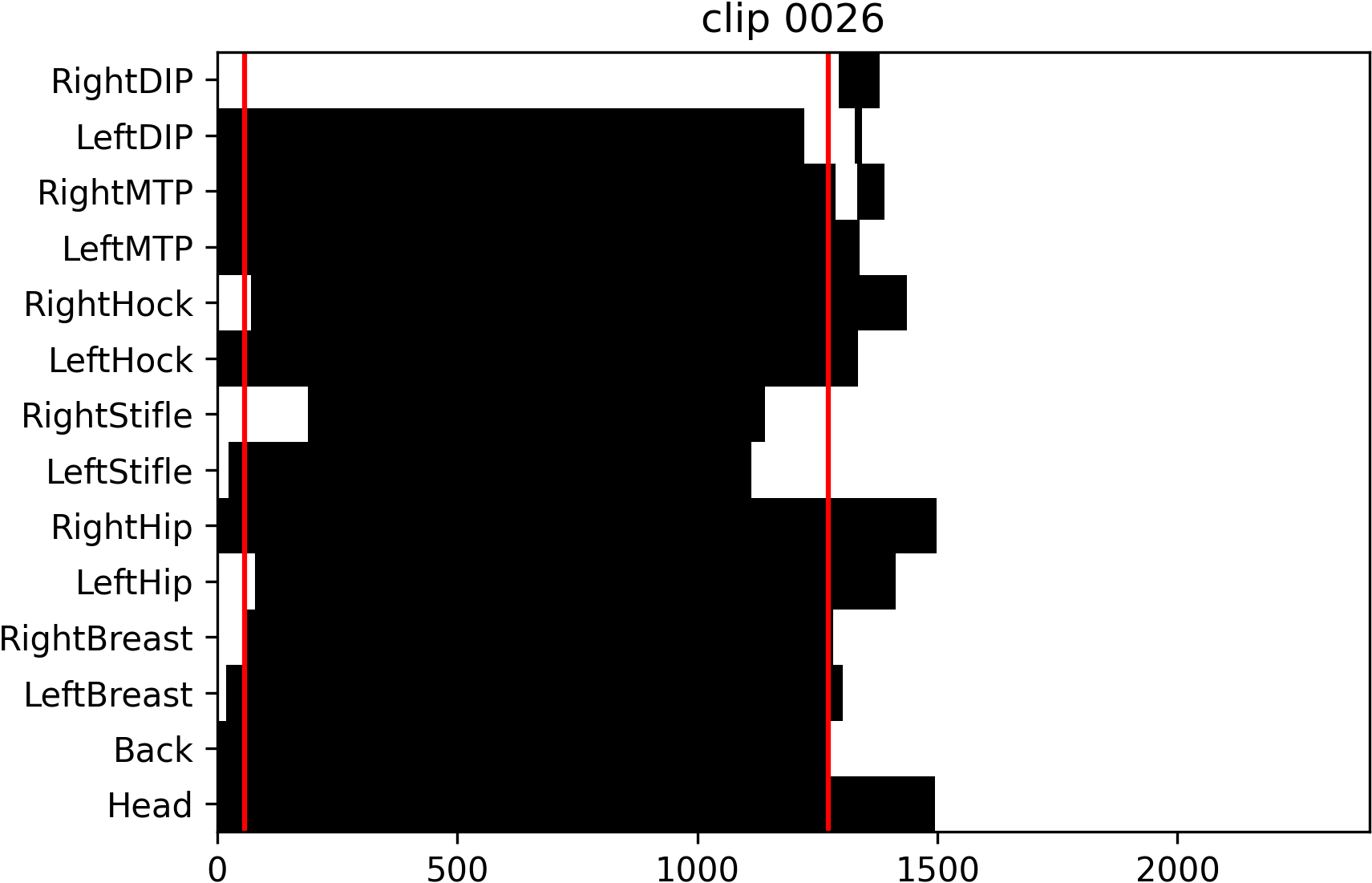}
\includegraphics[width=0.4\linewidth]{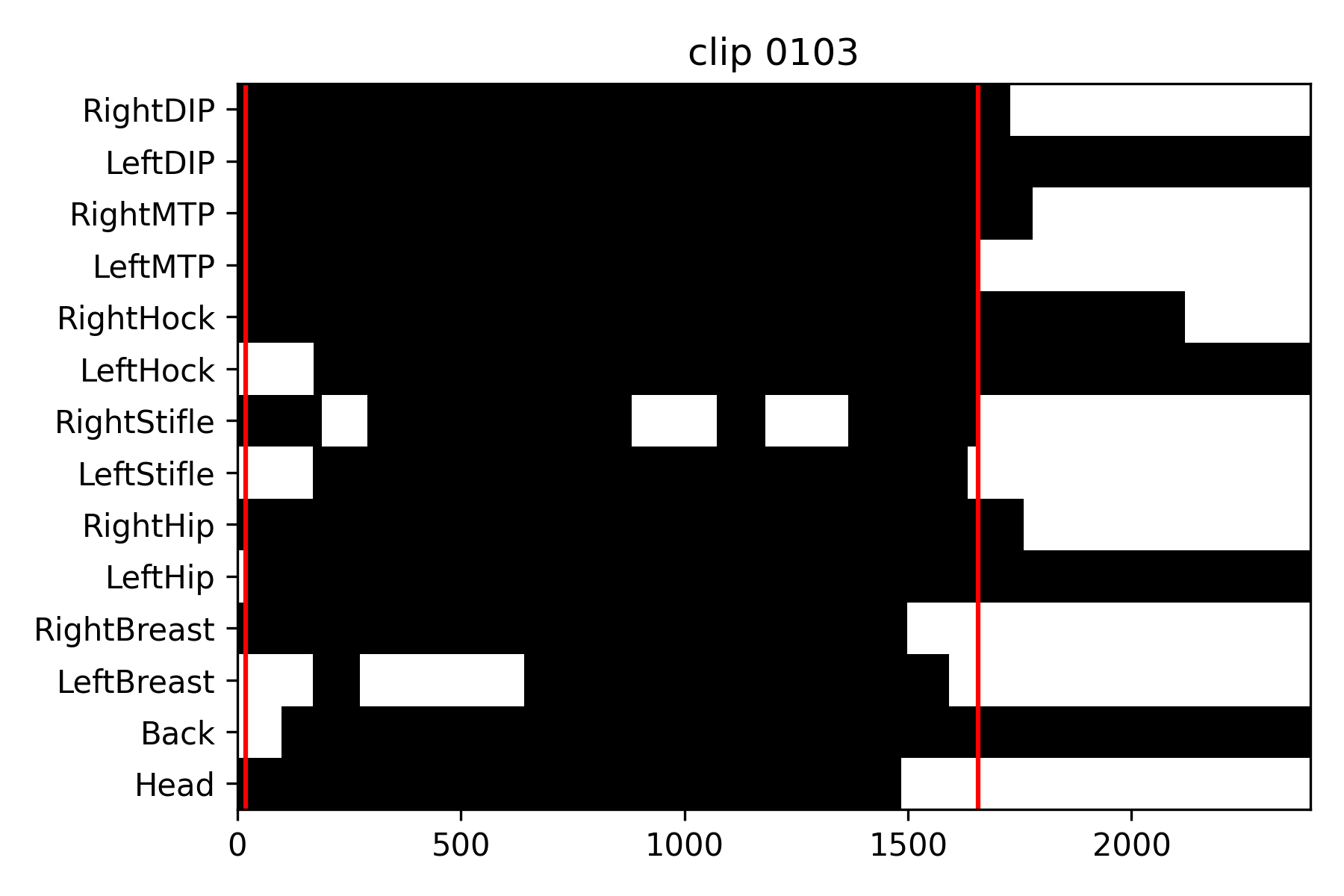}
\caption[Noisy motion capture data]{Binary masks showing, in black, the available data for the 14 markers, on 4 clips. The red lines indicate the intervals we used. The remaining missing markers had to be predicted.}
\label{fig:ostrichrl_original_mocap_data}
\end{figure}

\paragraph{Original mocap data}
We did not find any open-source motion capture database of ostrich movements, but after contacting the authors of \citet{jindrich2007mechanics}, we obtained the data originally used to study joint kinematics of ostriches performing cutting maneuvers. The provided dataset was composed of 82 clips, recorded at 240 Hz, containing the 3D coordinates of 14 markers distributed as follows: 1 the head, 1 on the spine, and 1 on each breast, hip, knee, ankle, mtp, and toe.

\begin{figure}[t]
\centering
\includegraphics[width=0.7\textwidth]{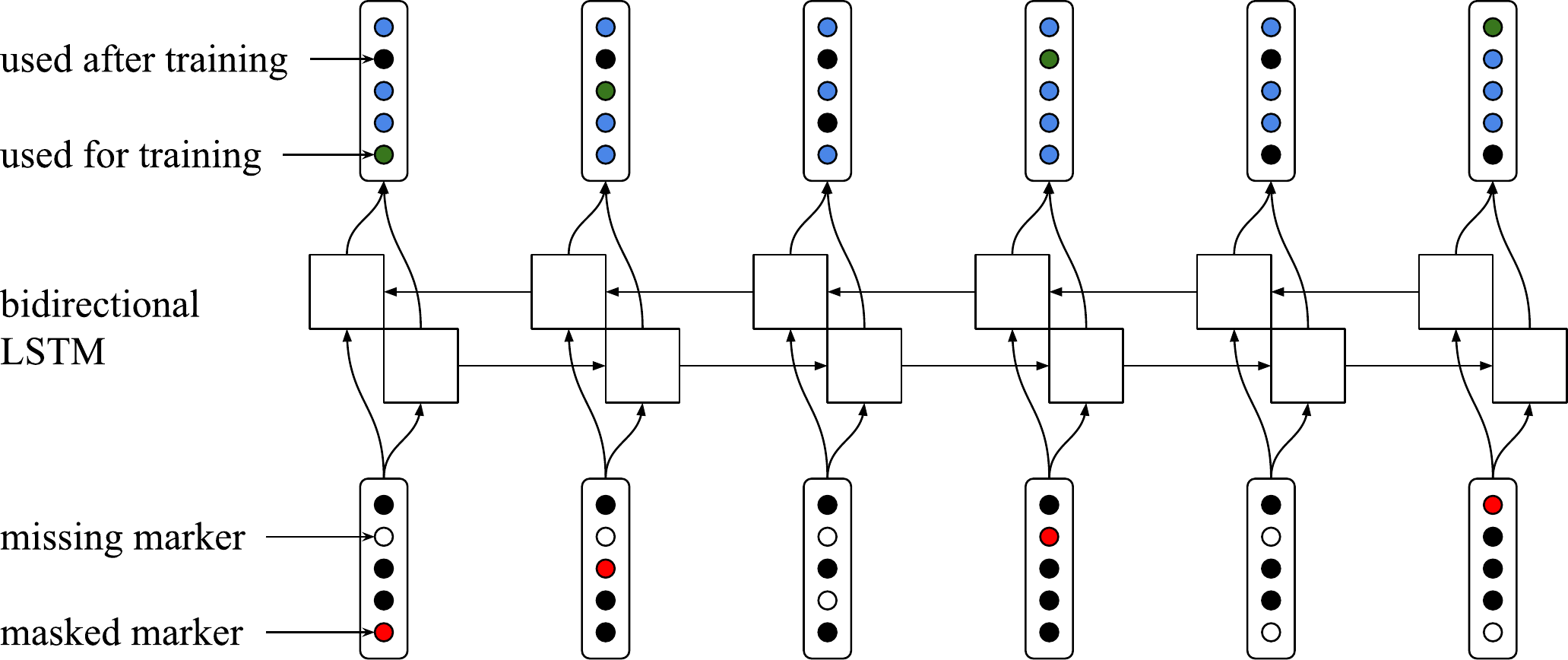}
\caption[Bidirectional LSTM used to predict the missing markers]{Illustration of the bidirectional LSTM used to predict the missing markers. The sequence of masked coordinates are encoded, then a sequence of decoded coordinates are produced, including predictions for the masked ones.}
\label{fig:ostrichrl_blstm}
\end{figure}

\begin{figure}[t]
\centering
\includegraphics[width=\textwidth]{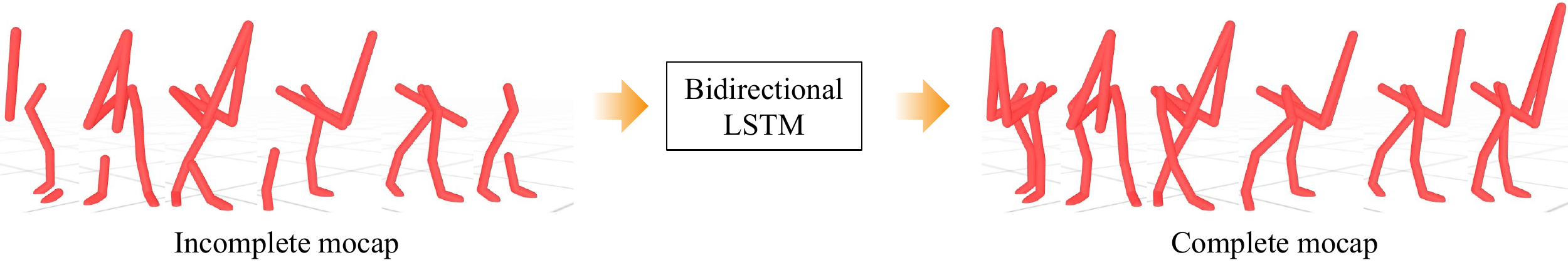}
\caption[Prediction of missing markers]{Visualization of the mocap prediction process. Segments are drawn between pairs of existing markers. Left: The original incomplete data. Right: The complete data after predicting the missing markers with a bidirectional LSTM.}
\label{fig:ostrichrl_mocap_generation}
\end{figure}

\paragraph{Cleaning of the mocap clips}
We selected the largest part of the clips where at least 10 markers were simultaneously present for at least 1 second, typically disregarding the beginning and the end of the clips when the ostrich was not in the field of view of the cameras. This selection left 35 clips. However, even in those clips, we found that a large number of markers were periodically missing, probably due to feathers and limbs occluding them from the cameras, as can be seen in \autoref{fig:ostrichrl_original_mocap_data}. To overcome this issue, we used a bidirectional LSTM trained in a similar way as a denoising autoencoder, as shown in \autoref{fig:ostrichrl_blstm}. With a probability of $0.1$, we randomly masked some markers and asked the model to predict their value given the context of a segment of $100$ steps from the same clip. We then used this model to predict the actual missing markers. Since the model was trained and then used on the same, relatively small dataset, we anticipated some overfitting issues, but playing the clips with the predicted markers in place of the missing ones gave surprisingly good results, as shown in \autoref{fig:ostrichrl_mocap_generation}.

\paragraph{Conversion to joint values}
After rescaling the marker coordinates to match the size of our model, the next step was to transform the set of marker coordinates to joint positions. Fortunately, enough markers were present in the different parts of the ostrich body to limit the space of solutions, except for the neck, where only one marker was present on the head. We used gradient descent with the mean Euclidean distance between the reference markers and those produced by joint poses on two sets of parameters simultaneously. The first set described where to attach the markers on the body parts and is shared across all the clips. This was needed because we did not have the exact location of the markers used during the motion capture acquisition and how they would translate to locations with respect to the bones. The second set of parameters described the joint values at every step. To reduce the space of solutions for the neck shape, we added a regularization term, encouraging the neck to follow an S shape. We first tried to use a finite difference approximation of the gradient, but found this approach to be too slow to be practical. We then decided to implement the kinematics function in a differentiable way, mapping joint values to body locations to create marker locations. This function was implemented using differentiable operations from the TensorFlow library \citep{abadi2016tensorflow}. We found this approach to be particularly fast thanks to the possibility to simultaneously optimise for all the steps across all the clips using the batch dimension. Finally, the velocity of the joints was inferred using the rate of change between two consecutive steps.

\paragraph{Cyclic clip}
We also created a cyclic running clip that can be used as a reference for running over an extended period of time. We used the middle section of one of the clips, which we repeated several times, and smoothed to remove discrepancies at the boundaries. This clip helped us produce the simulated biomechanical data for a cyclic gait that we compared to muscle excitations and lengths measured on emus in \autoref{subsec:ostrichrl_electromyography}.

\section{Experiments}

The proposed model was iteratively improved using a set of reinforcement learning tasks. The performance and visual quality of the solutions allowed us to select values for some muscle parameters such as muscle length ranges and forces, or to identify issues with unrealistic joint ranges. We focused on two types of tasks: locomotion and neck control. We found these to be particularly useful when designing the model, because they used the two main groups of muscles present in the model.

\subsection{Running forward}

\begin{figure}[t]
\centering
\includegraphics[height=4.8cm]{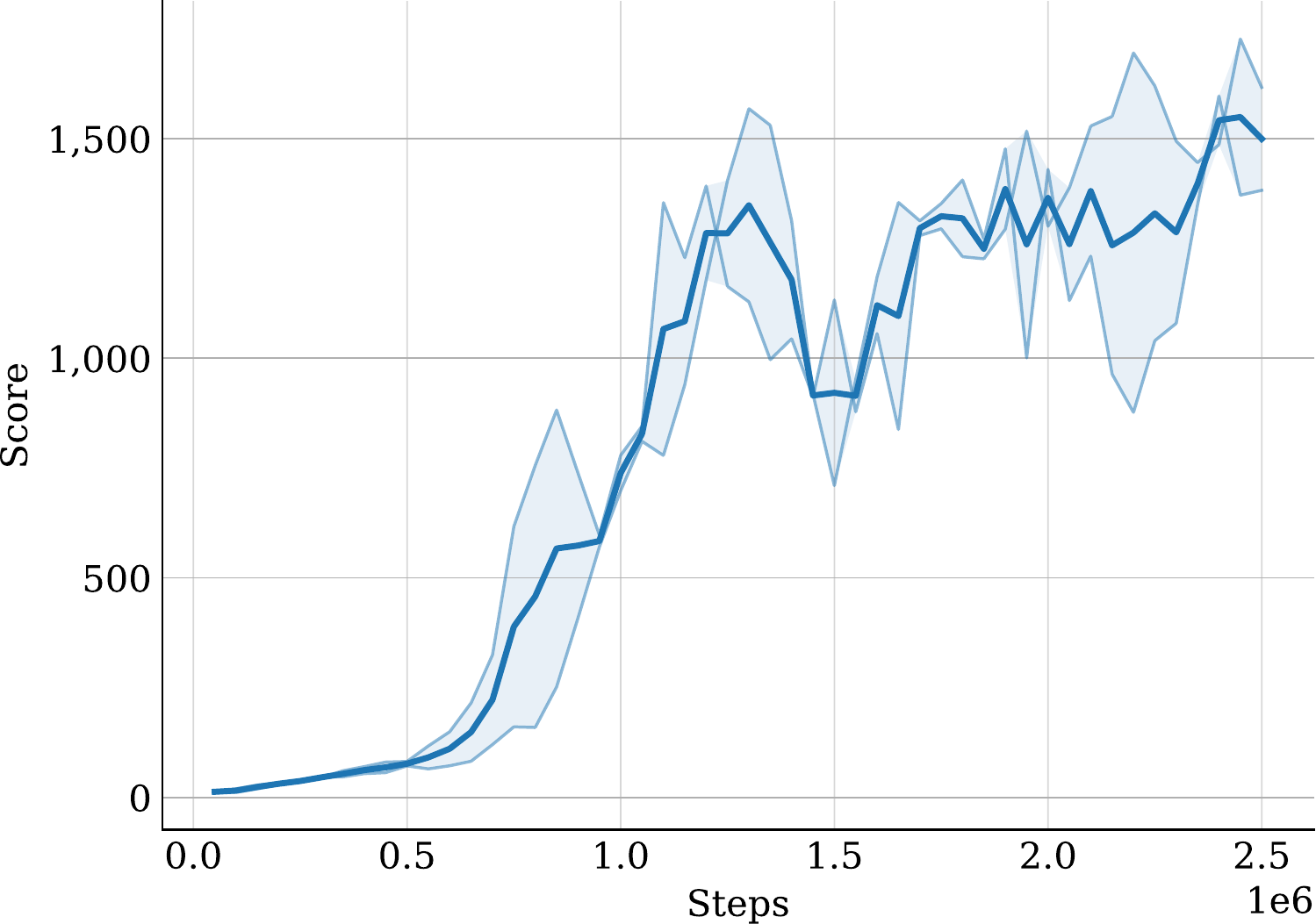}
\hfill
\includegraphics[height=4.8cm]{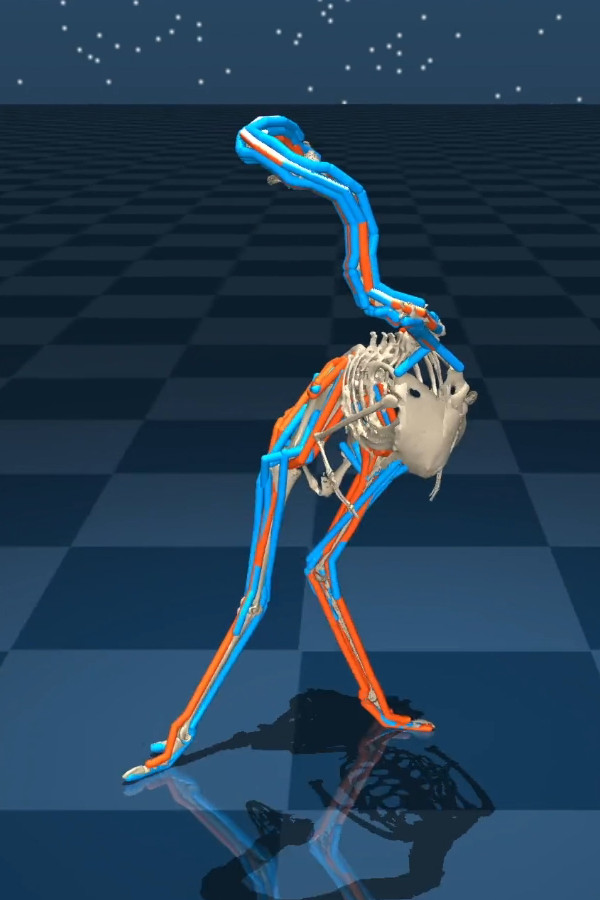}
\hfill
\includegraphics[height=4.8cm]{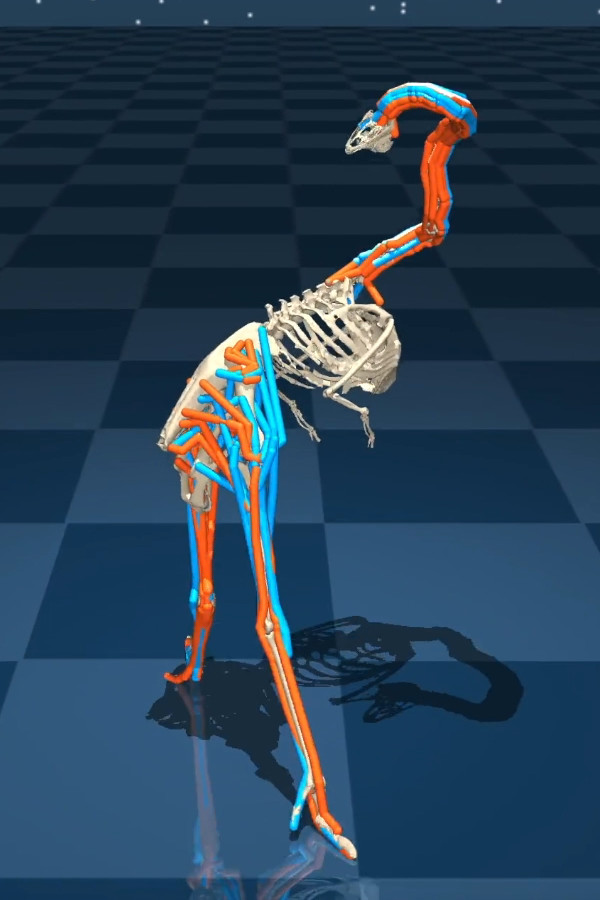}
\caption[Performance on run forward]{Performance on the ``run forward'' task. Sums of rewards are averaged over 2 seeds and smoothed with a sliding window of size 3. The agent manages to achieve satisfactory performance, but the gaits do not look natural as shown in the videos and images. Muscle excitations from -1 to 1 are represented by colors ranging from blue to orange respectively.}
\label{fig:ostrichrl_run}
\end{figure}

The first task we experimented with had a simple ``move forward'' objective, rewarding agents for running as fast as possible. It was initially constrained to a vertical planar space, it used torque-driven joints, and the neck was kept rigidly attached to the model's thorax. The goal was to ensure that the underlying model and simulation structure functioned as intended before adding muscles. The resulting gait was similar to that obtained when training on the 2D walker tasks of dm\_control and OpenAI Gym. After adding muscles to the legs, we found that the same gait repeatedly emerged: the toes were unrealistically bent backwards (plantarflexed) and the ankles remained straight. After a series of model adjustments on the different tasks, we were able to achieve more diverse and efficient policies on this one, even after relaxing the model constraint from 2D to full 3D space and actuating the neck. The final ``running forward'' task is described below, and the results can be seen in \autoref{fig:ostrichrl_run}.

\paragraph{Initialization}
The model is initialized in an upright pose with small random perturbations added to the leg and neck joints. Concretely, for these joints, the initial values are sampled uniformly in 1/5th of their respective intervals, centered around a default pose.

\paragraph{Observations}
The agent perceives the height of the head, pelvis, and feet, the joint positions except for the x-axis, the joint velocities, the muscle forces, activations, lengths and velocities, and the forward velocity of the center of mass on the x-axis.

\paragraph{Reward and termination}
The reward is simply the forward velocity of the center of mass on the x-axis. Episodes end if the head height falls below 0.9 meters, if the pelvis height falls below 0.6 meters, or if the torso rotation around the y-axis falls below -0.8 or exceeds 0.8 radians.

\subsection{Motion capture tracking}

The previous ``run forward'' task showed that learning to control such a complex articulated model with a simple reward function does not sufficiently regularize the search space, producing policies that are both unnatural and suboptimal. A typical solution to better constrain the policy space is to design a more sophisticated reward function. Reference motion tracking is a particularly well-suited option that has been used with motion capture data in many studies to produce natural-looking motions \citep{liu2010sampling,peng2018deepmimic,chentanez2018physics,merel2018neural,bergamin2019drecon,peng2019mcp,hasenclever2020comic}. As a proof of concept, we started with reference motion data that we labeled by hand, frame by frame, using videos of running ostriches found online. The task was initially planar and the model torque-driven. The quality of the results produced encouraged us to continue in this direction.

\begin{figure}[t]
\centering
\includegraphics[height=4.8cm]{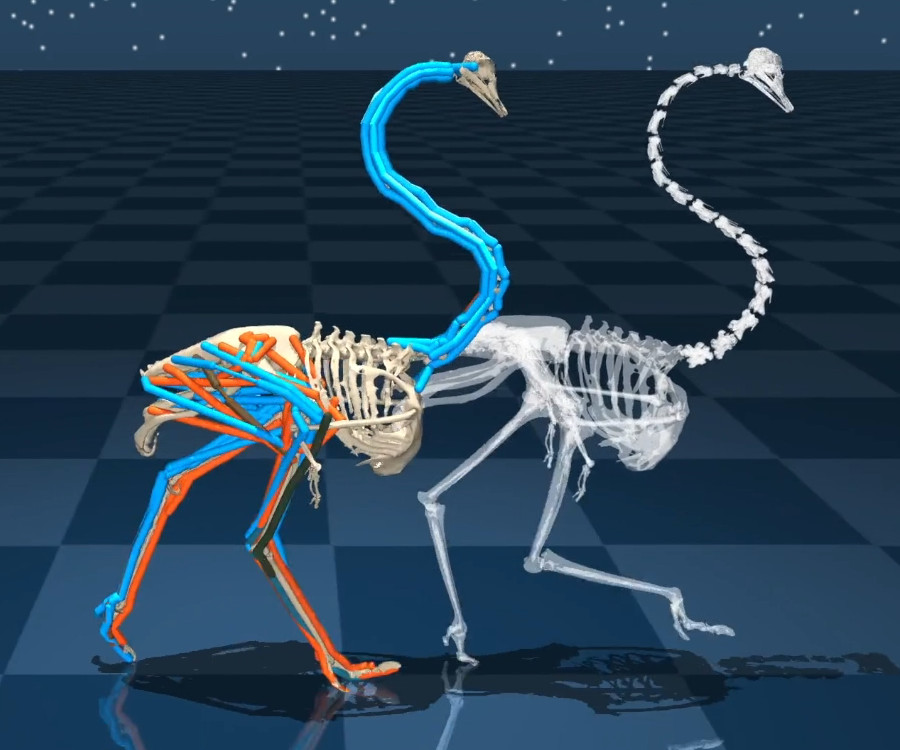}
\hspace{0.3cm}
\includegraphics[height=4.8cm]{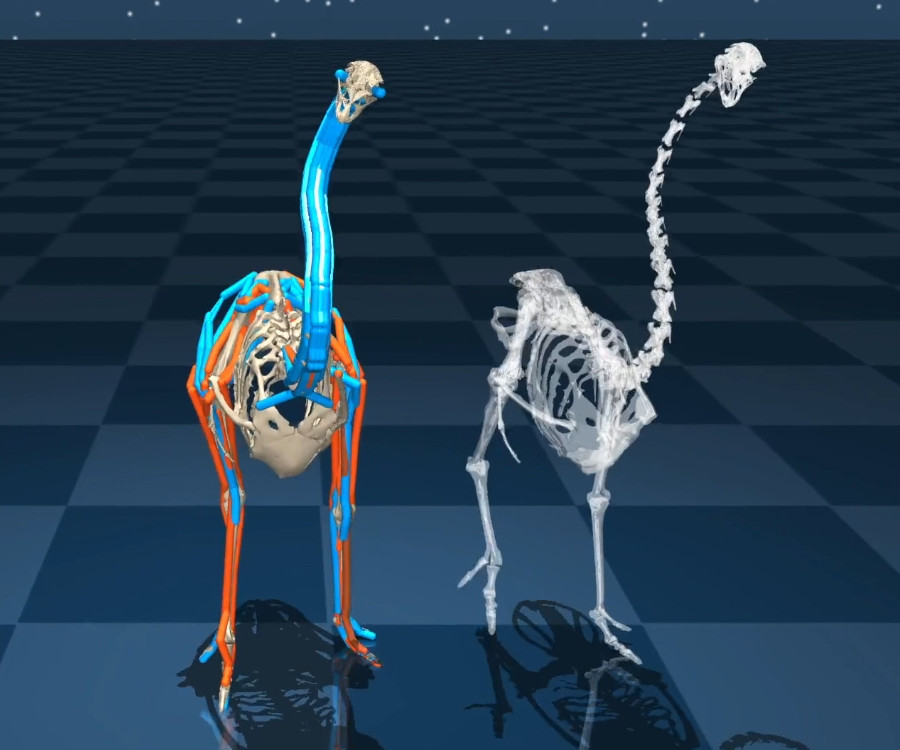}
\caption[Examples of mocap tracking]{Examples of motion capture tracking taken from two different clips. Left: A locomotion clip. Right: A complex clip requiring to stand still, perform neck movements, and turn in place. The reference is shown in white next to the model.}
\label{fig:ostrichrl_tracking}
\end{figure}

\begin{figure}[t!]
\centering
\includegraphics[width=\textwidth]{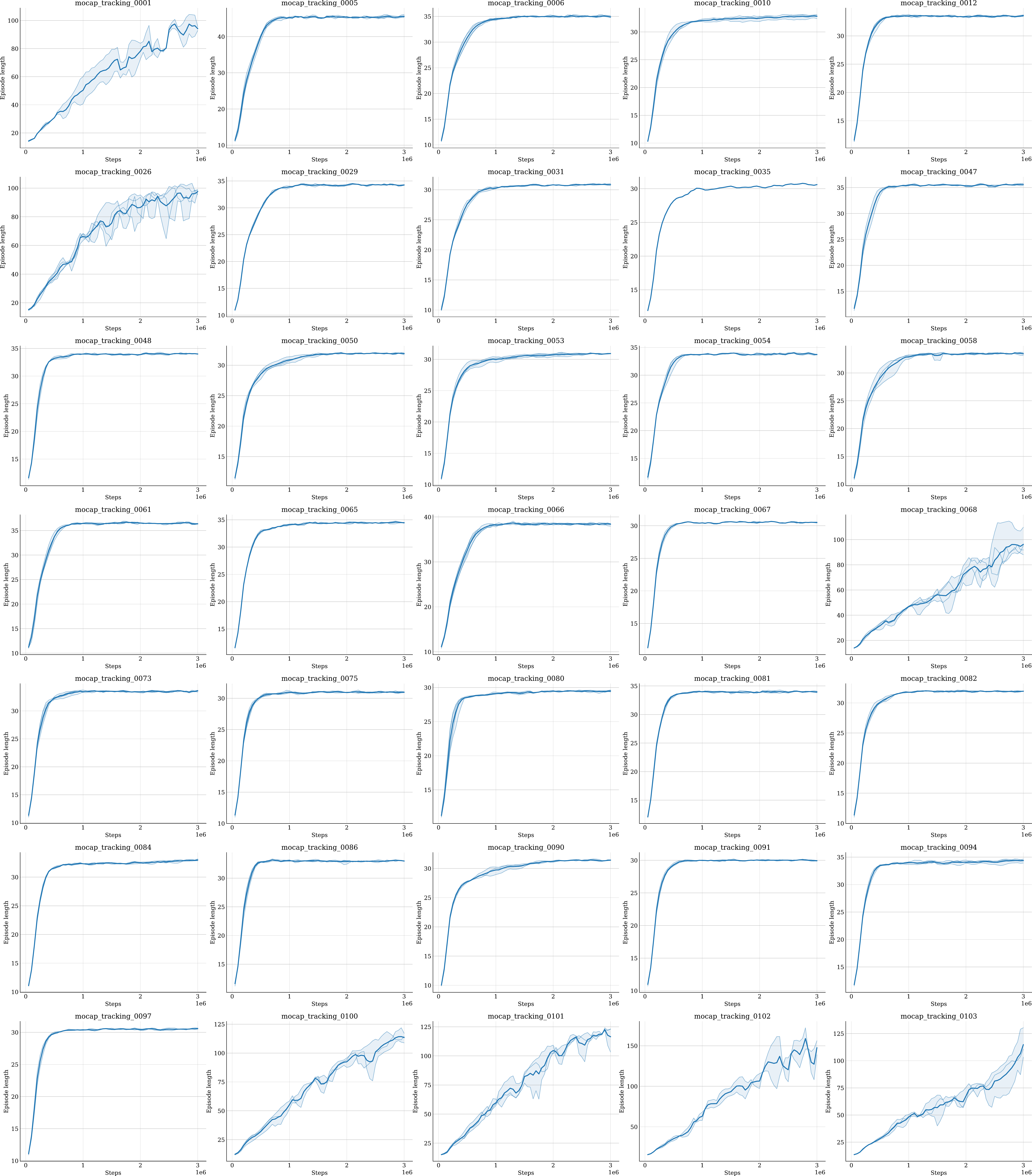}
\caption[Tracking performance on all clips]{Average episode length for all the clips. As agents perform better at tracking, the episodes last longer. Values are averaged over 5 seeds and smoothed with a sliding window of size 3.}
\label{fig:ostrichrl_results}
\end{figure}

After producing a satisfactory dataset, described in \autoref{subsec:ostrichrl_mocap}, we experimented with a number of variations for the task components. By performing large sweeps over the 35 clips and the cyclic one, we arrived at the mocap tracking task described below. Examples of tracking on two different clips is illustrated in \autoref{fig:ostrichrl_tracking}, the performance on every clip is shown in \autoref{fig:ostrichrl_results} and some of the videos demonstrate the quality of the tracking. The result on short walking clips is very satisfactory. These are the curves that quickly converge. For the most challenging clips, including standing and abrupt changes in speed, learning takes more time, and it seems that longer training would be needed.

\paragraph{Initialization and step}
During training, an initial time-step is uniformly sampled over the length of the episode, ignoring the last 20 steps. The model is initialized in the pose corresponding to this step, and a small amount of Gaussian noise, with a scale of 0.02, is added to help diversify experiences. The velocity of the joints was set to the one of the reference without modification. Since the control frequency used in our experiments was 40 Hz, the task used every sixth data point of motion capture data (240 Hz) during tracking.

\paragraph{Observations}
The agent perceives the height of the pelvis and feet, the joint positions and velocities, the muscle forces, activations, lengths and velocities, and the time left in the clip to allow time-dependent policies and to deal with the finite horizon, as described in \citet{pardo2018time}.

\paragraph{Reward and termination}
The tracking reward is defined as a product of Gaussian kernels $\exp (-w_p e_p) \times \exp (-w_r e_r)$ across all body parts. The quantity $e_p = \lVert \bar{p} - p \rVert$ measures the Euclidean distance between the coordinates of the center of mass $p$ and the reference $\bar{p}$, while the quantity $e_r = \arccos {(( \text{tr} (\bar{R} R^T ) - 1 ) / 2 )}$ measures the angle of the difference rotation between the orientation matrix of inertia $R$ and the corresponding reference $\bar{R}$. The weights $w_p$ and $w_r$ control the wideness of the Gaussians, accounting for the different magnitudes and importances. We used the values $0.2$ and $0.1$ respectively. The reward is bounded in (0, 1] and the multiplicative nature ensures that if one of the components is too far from the reference, the whole reward decays to $0$. Episodes end when the reward falls below $0.01$ or when the end of the clip is reached.

\subsection{Neck control}

The neck is composed of 17 vertebrae, which allows for a wide range of shapes. To achieve a particular head position and orientation, an infinite number of solutions exist, and during running, the weight and pose of the neck greatly influence balance. Controlling the neck with muscles that cross several vertebrae therefore requires a sophisticated policy.

\begin{figure}[t]
\centering
\includegraphics[height=4.8cm]{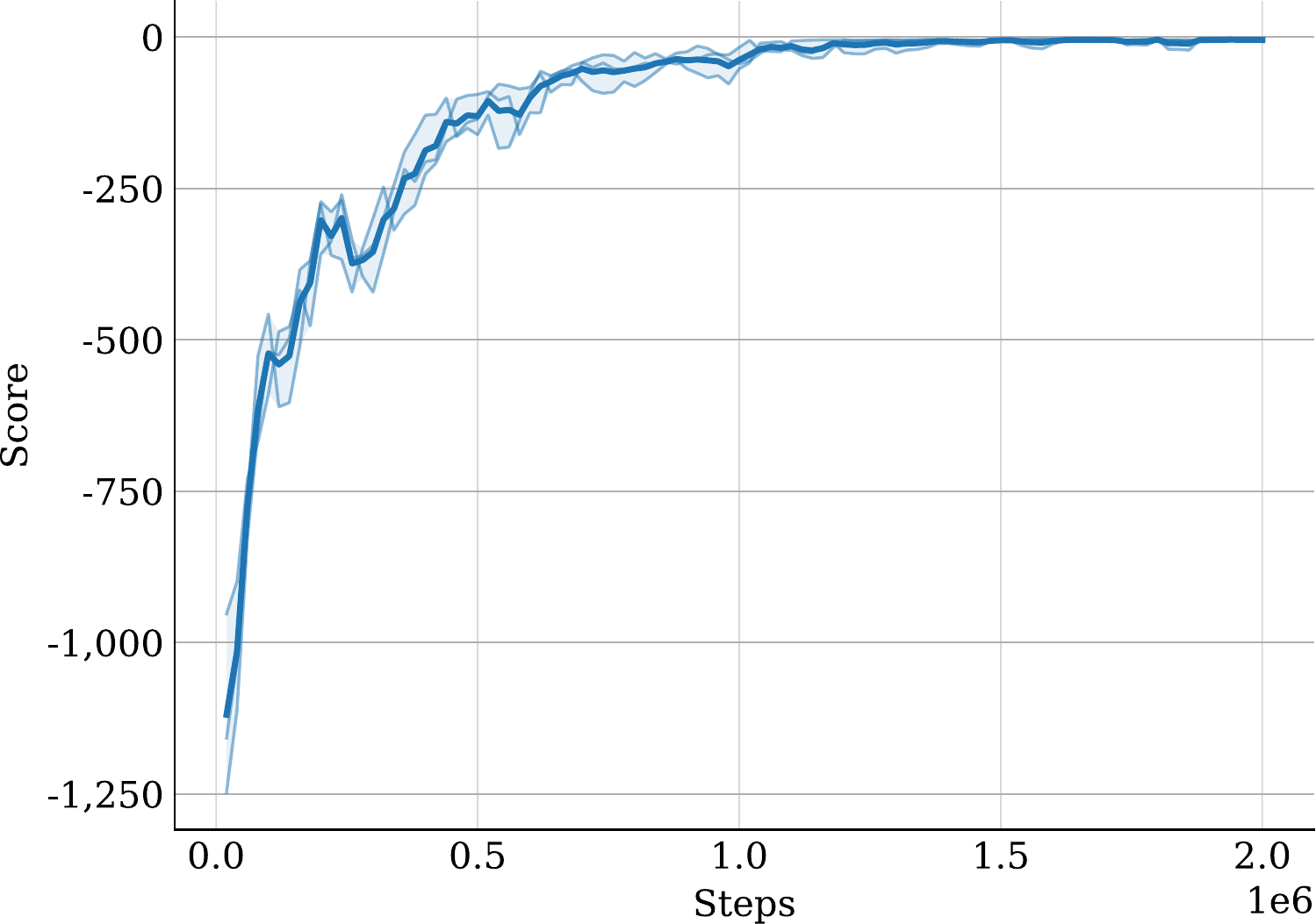}
\hfill
\includegraphics[height=4.8cm]{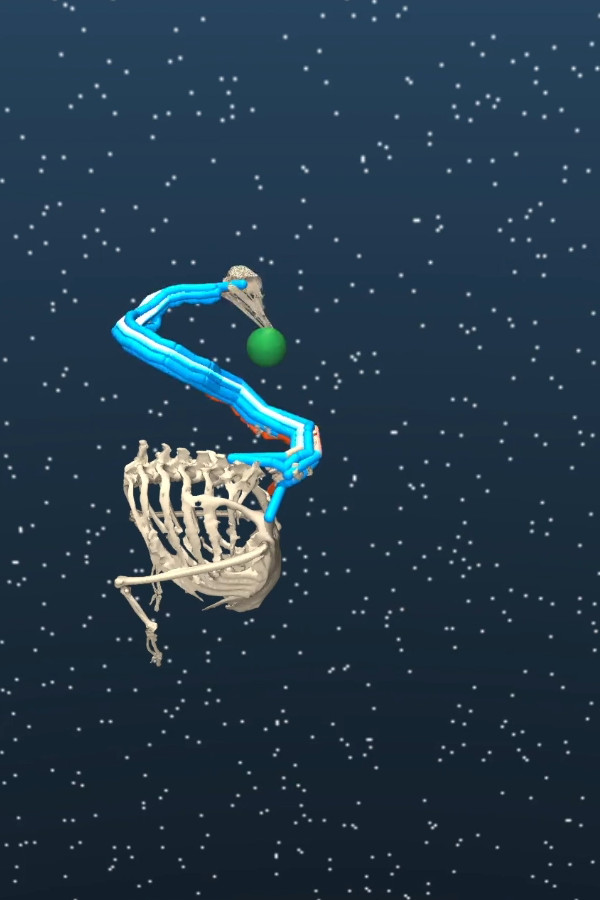}
\hfill
\includegraphics[height=4.8cm]{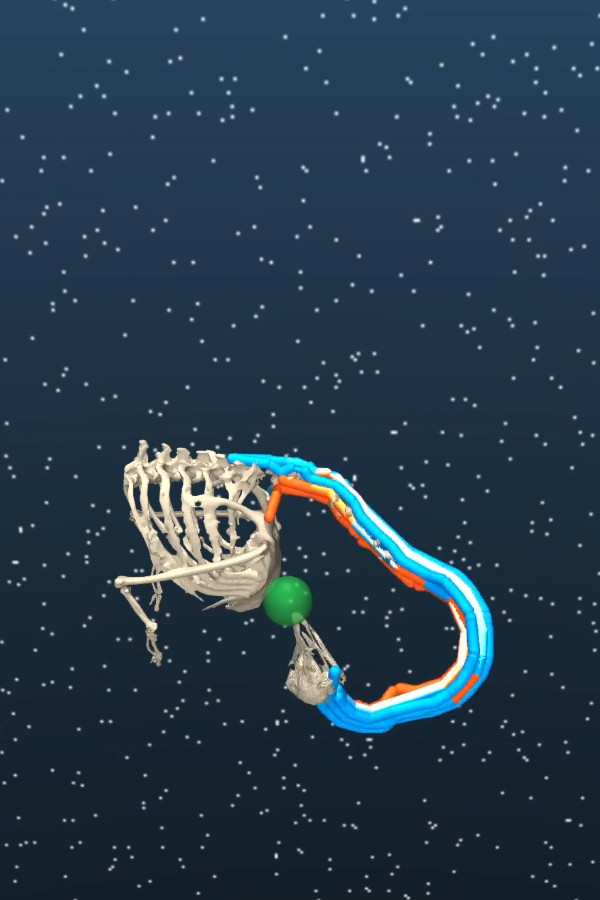}
\caption[Performance on neck control]{Performance on the ``neck control'' task. Sums of rewards are averaged over 3 seeds and smoothed with a sliding window of size 3. The agent manages to achieve satisfactory performance, and the neck shapes are fairly natural.}
\label{fig:ostrichrl_neck_results}
\end{figure}

To tune the neck portion of the model, we found that mocap tracking of the whole body was not ideal due to the large number of moving parts and the fact that the reference neck poses were artificial. Therefore, we created another task whose objective was to reach random targets with the beak. Only the rib cage, neck, and head were used and the rib cage was held fixed in space. Although effective policies were easily obtained, the initial neck shapes were very irregular, with many abrupt and unnatural changes. To reduce this phenomenon, we increased the stiffness of the neck joints until satisfactory smoothness was achieved. For the joint boundaries, we initially used the values of \citet{dzemski2007flexibility}, but then realized that the boundaries needed to be increased to allow for tighter neck curves that ostriches can perform. The final ``neck control'' task is described below, and the results can be seen in \autoref{fig:ostrichrl_neck_results}.

\paragraph{Initialization}
Every 100 episodes, and for the very first one, the neck is initialized with a default S shape. For all other episodes, the neck position and velocity are maintained from the last state in the previous episode. The target position is randomized to a feasible area. To do this, we use rejection sampling. Points are repeatedly sampled within a sphere centered at the base of the neck with a radius of 0.8 meters, slightly smaller than the length of the neck, and discarded when inside a second smaller sphere representing approximately the ostrich torso, with a radius of 0.6 meters.

\paragraph{Observations}
The agent perceives the joint positions and velocities, the muscle forces, activations, lengths and velocities, the coordinates of the beak and target, and the vector from the beak to the target.

\paragraph{Reward and termination}
The reward is simply the negative of the Euclidean distance between the beak and the target. When this distance falls below a threshold of 0.05 meters, episodes terminate.

\subsection{Experimental details}

To obtain the policies, we used a TD4 agent, proposed in \citet{pardo2020tonic}, a mixture of TD3 \citep{fujimoto2018addressing}, with its pair of critics, delayed actor training and target action noise, and D4PG \citep{barth2018distributed}, with its distributional value function \citep{bellemare2017distributional} and n-step returns. We also found that Ornstein Uhlenbeck exploration noise \citep{lillicrap2015continuous} was significantly better than Gaussian noise, probably due to the importance of correlated excitation when controlling muscles. We chose to use the Tonic deep RL library, because it provided us with a state of the art agent and the flexibility we needed to quickly try, evaluate and visualize experiments while allowing custom dm\_control tasks to be used. The different tasks can be visualized on the website.

For the experiments reported above, we used the following hyper parameters. The models are simple MLPs with 2 hidden layers of size 256, ReLU activations, and 51 atoms for the distributional heads. The replay buffer has a size of 1,000,000, batches of size 50 are sampled after the first 50,000 steps and then 50 times every 50 steps. We use 1-step returns, a discount factor of 0.99, a learning rate of 0.0001, a target action noise scale of 0.25, an Ornstein-Uhlenbeck action noise with scale 0.25 and 10,000 initial random steps. 

\subsection{Electromyography comparison}
\label{subsec:ostrichrl_electromyography}

\begin{figure}[t]
\centering
\includegraphics[width=\linewidth]{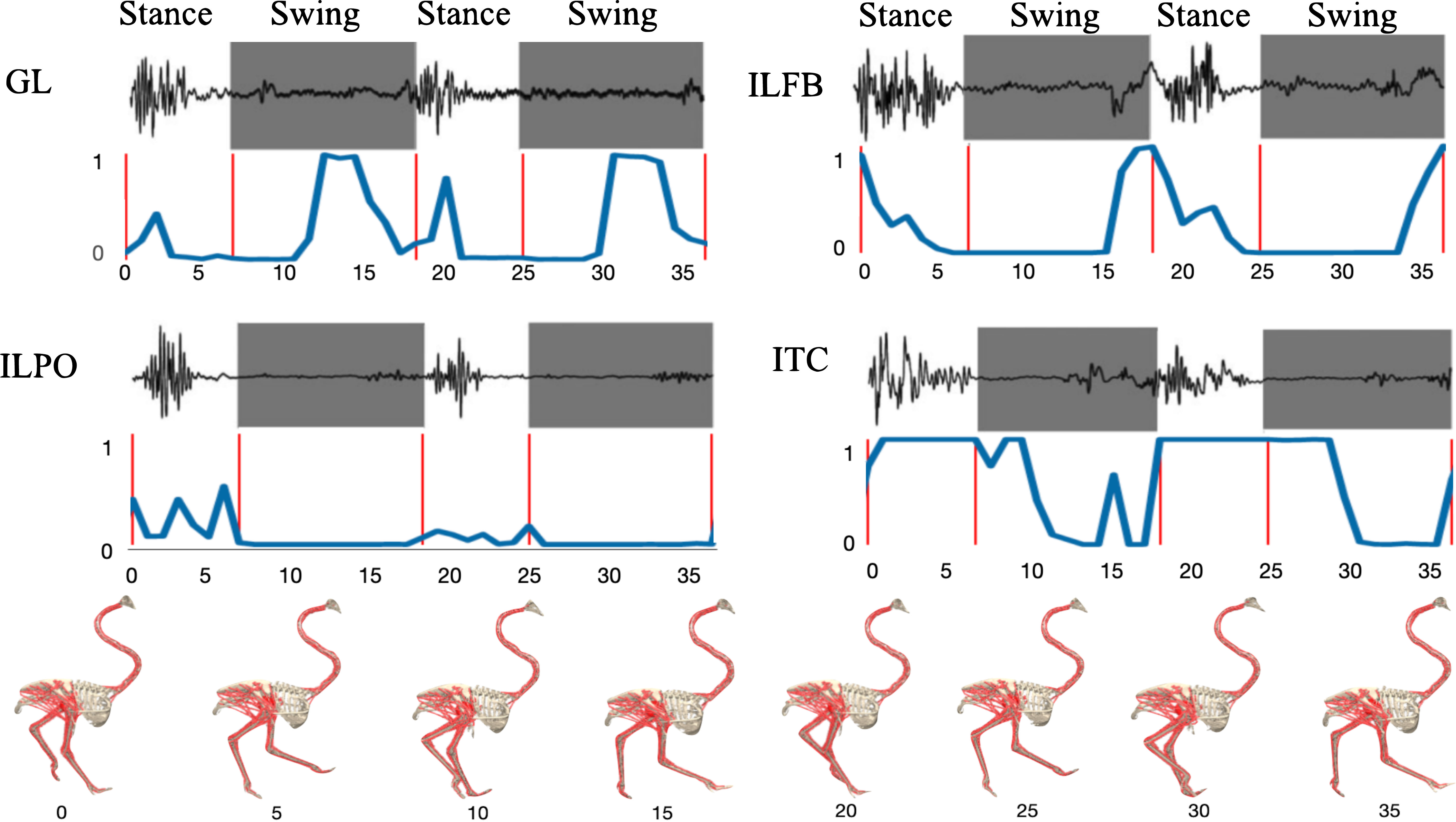}
\caption[Comparison with experimental EMG data from emus]{Comparison with experimental EMG data from emus. The first red vertical line indicates that the right foot was in contact with the ground (start of stance phase). The second one is when the foot leaves the ground (start of swing phase). EMG data from \citet{cuff2019relating} are on top. Muscle excitations (ours) are below.}
\label{fig:ostrichrl_emg}
\end{figure}

To demonstrate the realism of the proposed simulation, we chose to compare the muscle excitations outputted by a policy trained on the cyclic mocap tracking task, to electromyography (EMG) data. On one side, EMG measures electrical activity in muscles, in response to a stimulation from motoneurons and on the other side, raw actions produced by a policy are used by MuJoCo to produce muscle activations, as explained in \autoref{subsec:ostrichrl_muscle_dynamics}.

Collecting EMG data in animals is challenging because their skin may be too thick for the electrodes and surgery may be required to implant them directly into the muscles. Because no ostrich EMG data have yet been collected, we decided to perform comparisons with data from emus, close relatives of ostriches. EMG data were originally acquired by \citet{cuff2019relating} to study muscle recruitment patterns in different bird species and understand neuromuscular control from an evolutionary perspective. Their results showed that walking/running birds generally use their hind limb muscles in very similar ways, making it reasonable to compare emu EMG data with ostrich simulations.

To compare the data, following standard conventions in locomotion research, we divide the gait into two phases: stance (foot is on the ground) and swing (foot is off the ground). Once the gait has been segmented into these phases, we scale them accordingly between the two studies, to account for the difference in speed. As show in \autoref{fig:ostrichrl_emg}, we find broadly similar excitation patterns, except for the GL (gastrocnemius lateralis) which had an extra burst of swing phase excitation in these simulations, not usually found in any extant birds \citep{cuff2019relating}. The ITC (iliotrochantericus caudalis) also had simulated excitation throughout more of the swing phase than in the EMG data, reminiscent of findings for simulated ostriches using the OpenSim model in \citet{rankin2016inferring}. There are many more muscles than actual degrees of freedom in the legs, allowing many possible excitation patterns to match the kinematic data. In addition, because unnecessary excitations are not penalized, it is expected to see more excitations than in the experimental data.

\subsection{Compatibility with Cassie}

Cassie is a bipedal robot produced by Agility Robotics. It has undergone extensive work in trajectory optimization \citep{reher2019dynamic,li2020animated,apgar2018fast} and reinforcement learning \citep{xie2018feedback,xie2020learning,li2021reinforcement}. Its morphology is similar to that of ostrich legs with short thighs, however, because Cassie's joints are powered by electric motors, they are each limited to one degree of freedom. To create multiple degrees of freedom, several joints are added in series and an offset is necessary between them for mechanical reasons. This is a major difference between the ball-and-socket joints allowed by musculoskeletal bodies and traditional robots. In addition, Cassie couples the knee and ankle joints using rods. However, the similarity to the ostrich morphology seemed sufficient to attempt to apply some of our tasks to a model of this robot. This provides an interesting opportunity to compare muscle and torque actuation with two different real-world body models with fairly similar morphologies.

\begin{figure}[t]
\centering
\includegraphics[width=3.2cm]{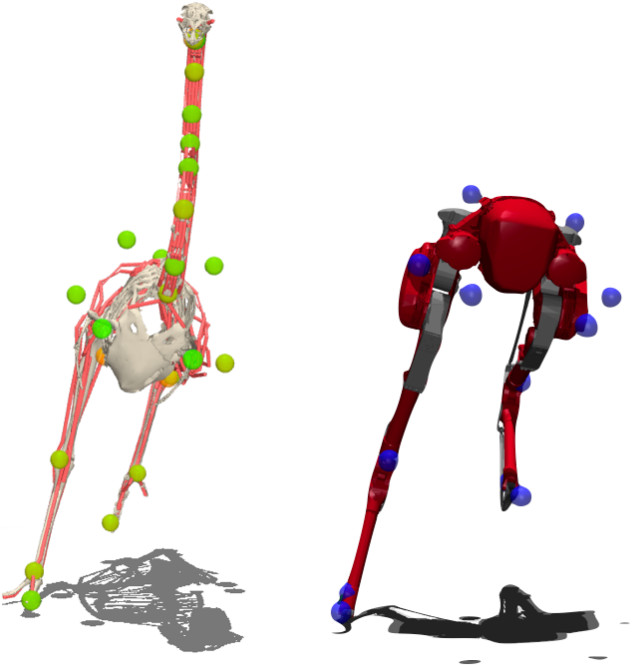}
\hfill
\includegraphics[width=10.0cm]{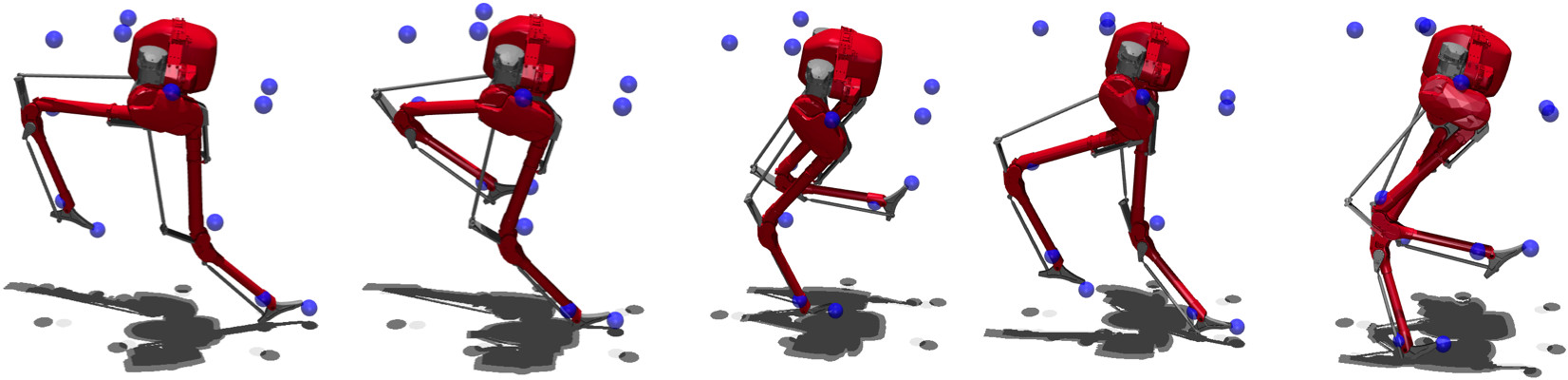}
\caption[Motion capture applied to Cassie]{Ostrich and Cassie markers comparison, and poses from a motion capture clip. The morphology of the two models is similar enough to allow the motion capture data to be reused.}
\label{fig:ostrichrl_cassie}
\end{figure}

We started with Cassie's MuJoCo model, provided by the Oregon State University Dynamic Robotics Laboratory \footnote{\url{https://github.com/osudrl/cassie-mujoco-sim}}. The model had to be adjusted to increase its stability and some equality constraints had to be added to ensure that the parts were properly connected with the rods. In order to obtain the robot poses corresponding to the mocap clips, except for the neck, we used the mocap generation pipeline described in \autoref{subsec:ostrichrl_mocap}. The only modification was the use of MuJoCo's Jacobian function during the gradient descent step \citep{buss2004introduction}, to satisfy the equality constraints. Surprisingly, Cassie's morphology is close enough to that of an ostrich that the clips can be reproduced fairly accurately, as shown in \autoref{fig:ostrichrl_cassie}. The resulting dataset provides an interesting set of behaviors achievable with the Cassie robot, including steps that are more natural than those typically found in the literature and demos using this robot.

\begin{figure}[t]
\centering
\includegraphics[height=4.8cm]{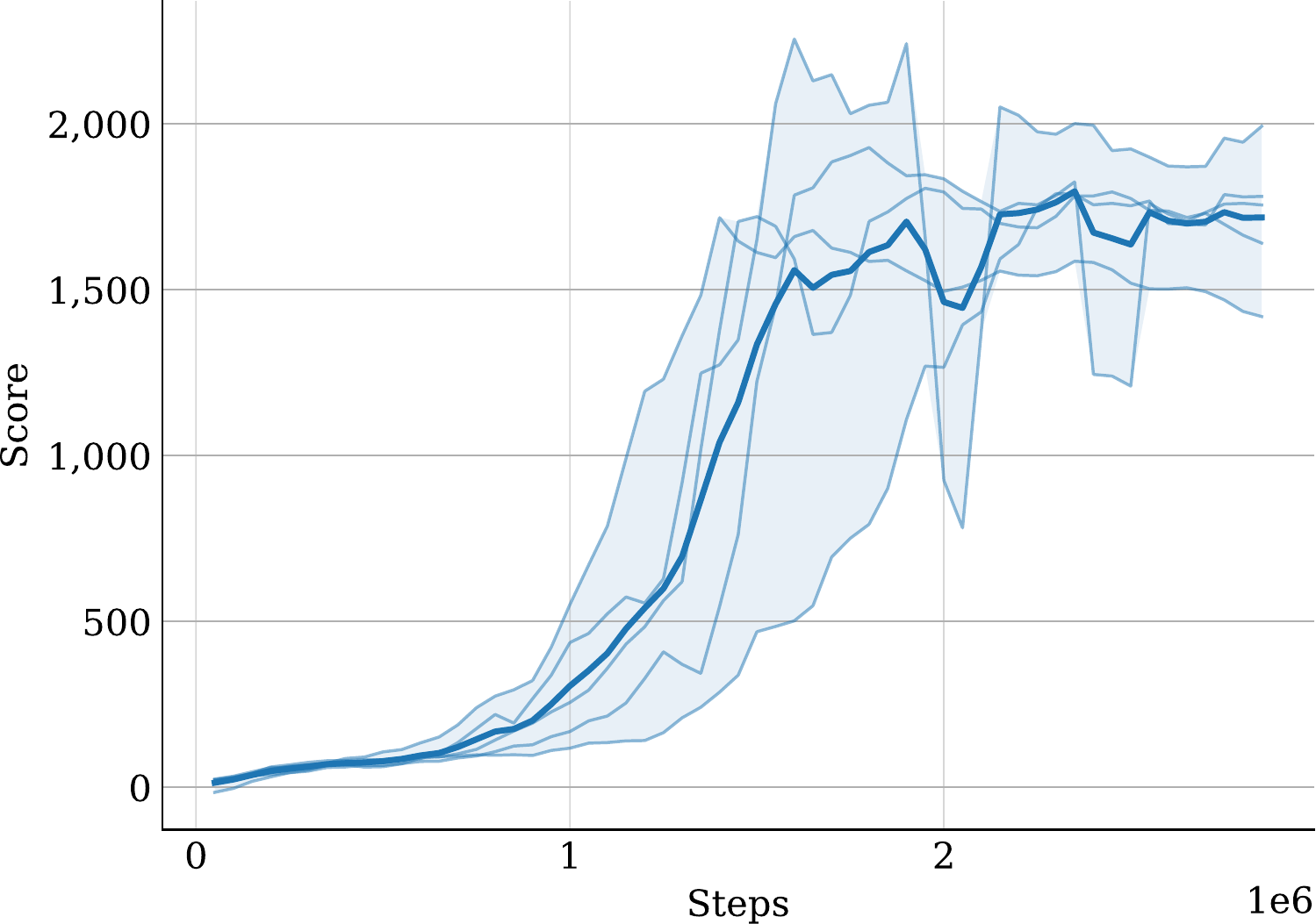}
\hfill
\includegraphics[height=4.8cm]{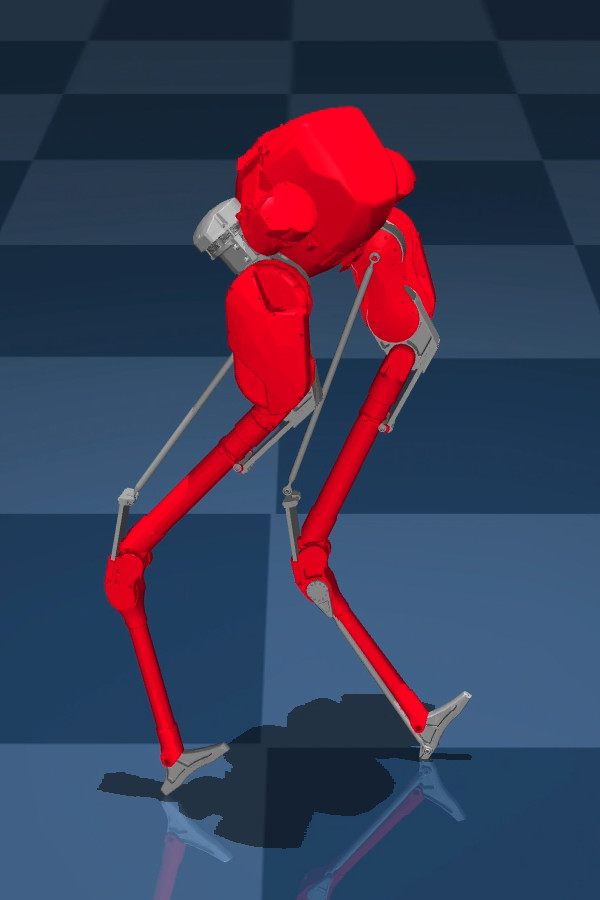}
\hfill
\includegraphics[height=4.8cm]{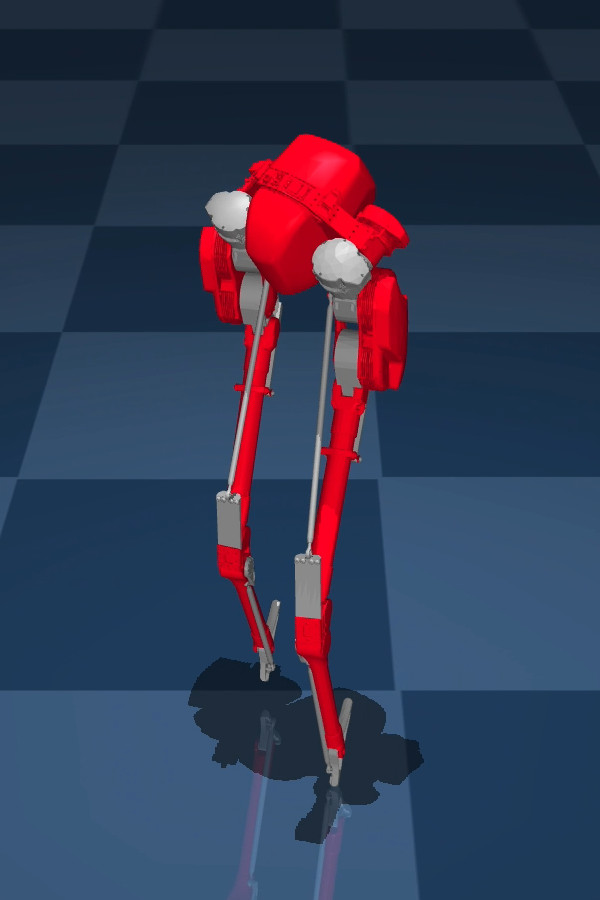}
\caption[Performance on run forward with Cassie]{Performance on the ``run forward'' task with the Cassie model. Sums of rewards are averaged over 5 seeds and smoothed with a sliding window of size 3. The agent manages to achieve a slightly faster speed than with the musculoskeletal ostrich model.}
\label{fig:ostrichrl_run_cassie}
\end{figure}

Unfortunately, Cassie's model remains quite unstable on the motion capture tracking task, probably due to constraints in the rods and more work would be needed to achieve a satisfactory level of tracking on all clips. However, we also adapted the ``run forward'' task with more success, as shown in \autoref{fig:ostrichrl_run_cassie}.

\section{Discussion}

We presented a new musculoskeletal model of an ostrich. This is the only existing complete model of the fastest biped on Earth, an appropriate choice for studying fast locomotion. The proposed MuJoCo simulation is significantly faster than most available musculoskeletal models, usually implemented with OpenSim. Anatomical data were used to ensure that the model was reasonably realistic. We also released a motion capture dataset of ostrich behaviors and open-sourced a set of reinforcement learning tasks built with dm\_control. The neck control task is novel and specific to the unique morphology and control aspects of the model. The motion capture tracking task, while similar to those found in the literature, has a unique tracking objective based on matching body part locations and rotations instead of relying on joints and markers. We found that this method produced better tracking results.

The unique combination of a fast and realistic musculoskeletal model with reinforcement learning tasks provides an interesting new benchmark for RL agents in general. Muscle-actuated systems are very different from torque-actuated systems and present specific challenges that we believe could be of interest to the RL community. These include over-actuation, filtered temporal dynamics, and nonlinear force curves. In addition, it is well known that when learning with a simple reward function to control complex articulated models, policies tend to produce odd and suboptimal behavior. In effect, we found that, compared to motion capture tracking, the running task produced very unnatural and inefficient policies. A major challenge for the RL community could therefore be to try to improve the policies discovered on tasks that do not rely on motion capture data. Research directions could involve more efficient exploration methods, new inductive biases in neural architectures, or other types of regularizations, including energy efficiency and fatigue \citep{potvin2017motor}. We believe that our model and task set are ideal for this type of research. In addition, new tasks can easily be created to study different problems, such as jumping or adapting to rough terrain, and we would be happy to support their inclusion in the repository.

Furthermore, this work is an interdisciplinary effort, with the goal of bridging several research communities. We believe that the interaction between fields such as biomechanics, reinforcement learning, graphics, robotics, and computational neuroscience can produce new understanding of biological bodies and advances in artificial ones. The use of reinforcement learning to improve our model can be more widely accepted for biomechanics research in place of more traditional techniques and can help produce synthetic muscle activation data. This data can in turn help fields such as sports science to create more effective training or recovery programs \citep{coste2017comparison}. In addition, the computer graphics community is striving for more true-to-life animations, and musculoskeletal simulations could be an important asset in producing more natural movements while introducing more accurate physical interactions and skin deformations when using volumetric muscles. Roboticists have attempted to implement robots powered by artificial muscles, and our work could be used to perform comparisons in energy efficiency and joint movement capabilities. Finally, in computational neuroscience research, models of motor neurons, motor primitives, and muscle synergies could be tested using our simulation.

\section*{Acknowledgements}
The research presented in this paper has been supported by the Royal Veterinary College and Dyson Technology Ltd.

\bibliography{main}
\bibliographystyle{iclr_2021}

\end{document}